\documentclass[]{interact}

\usepackage{url}
\usepackage{subfigure}
\usepackage{epstopdf}
\usepackage[caption=false]{subfig}

\usepackage{natbib}
\bibpunct[,
]{(}{)}{;}{a}{}{,}
\renewcommand{\bibfont}{\fontsize{10}{12}\selectfont}

\theoremstyle{plain}

\theoremstyle{definition}

\theoremstyle{remark}

\begin{document}
    \title{Positioning radiata pine branches requiring pruning by drone stereo
    vision}

    \author{\name{Yida Lin\textsuperscript{a}, Bing Xue\textsuperscript{a}, Mengjie Zhang\textsuperscript{a}, Sam Schofield\textsuperscript{b}, Richard Green\textsuperscript{b}}
    \affil{\textsuperscript{a} Centre of Data Science and Artificial Intelligence $\&$ School of Engineering and Computer Science, Victoria University of Wellington, PO Box 600, Wellington, New Zealand}
    \affil{\textsuperscript{b} Computer Vision Research Lab $\&$ Department of Computer Science and Software Engineering, Canterbury University, PO Box 8041, Christchurch, New Zealand}}

    \maketitle
    \begin{abstract}
        This paper presents a stereo-vision-based system mounted on a drone for detecting
        and localising radiata pine branches to support autonomous pruning. The
        proposed pipeline comprises two stages: branch segmentation and depth estimation.
        For segmentation, YOLOv8, YOLOv9, and Mask~R-CNN variants are compared on
        a custom dataset of 71 stereo image pairs captured with a ZED Mini camera.
        For depth estimation, both a traditional method (SGBM with WLS filtering)
        and deep-learning-based methods (PSMNet, ACVNet, GWCNet, MobileStereoNet,
        RAFT-Stereo, and NeRF-Supervised Deep Stereo) are evaluated. A centroid-based
        triangulation algorithm with MAD outlier rejection is proposed to compute
        branch distance from the segmentation mask and disparity map.
        Qualitative evaluation at distances of 1--2\,m indicates that the deep learning-based
        disparity maps produce more coherent depth estimates than SGBM,
        demonstrating the feasibility of low-cost stereo vision for automated
        branch positioning in forestry.
    \end{abstract}


    \section{Introduction}

    Pinus radiata, commonly known as radiata pine, is a highly valuable species
    extensively cultivated in New Zealand due to its rapid growth and versatile
    applications in forestry and timber industries. This species is essential for
    producing high-quality timber used in construction, paper manufacturing, and
    other wood-based products, significantly contributing to the economy
    \cite{lin2024drone,mason2023impacts}. For instance, in New South Wales, Australia,
    the radiata pine industry contributed approximately \$3 billion to the economy
    in 2021-22, highlighting its economic importance\footnote{https://www.dpi.nsw.gov.au/dpi/climate/climate-vulnerability-assessment/forestry/radiata-pine}.
    However, to ensure the trees grow with strong, straight trunks and produce
    clear wood free of knots, regular pruning is necessary. Traditionally
    performed manually, tree pruning and trimming are hazardous occupations globally,
    posing significant challenges and dangers. According to Tree Care Industry Magazine\footnote{https://tcimag.tcia.org/safety/tree-work-safety-by-the-numbers/},
    in the United States alone, the Bureau of Labor Statistics reports a
    fatality rate of 110 per 100,000 tree trimmers and pruners, which is about 30
    times higher than the average across all industries. Moreover, non-fatal injury
    rates for tree workers are also substantially higher, at approximately 239
    injuries per 10,000 workers, compared to 89 per 10,000 across all industries.
    The hazardous nature of the work also contributes to persistent labour
    shortages in the sector.

    The application of drones has significantly expanded due to technological
    advancements, yet their use in forestry, particularly for branch pruning, remains
    limited. Existing drone systems designed for pruning often require manual
    operation and can only cut thicker branches, relying on costly auxiliary equipment
    like LiDAR sensors, which hinders their widespread adoption
    \cite{molina2017aerial}. To address these challenges, we aim to develop a
    drone equipped with a stereo camera and a pruning tool, capable of automatically
    detecting and pruning branches as thin as 10 mm. This system utilizes the
    stereo camera for branch identification and distance measurement, enabling autonomous
    pruning. By streamlining the design and enhancing sensor technology, this approach
    aims to make drone-based pruning more accessible and cost-effective,
    improving precision and efficiency in forestry management.

\begin{figure}[htbp]
        \centering
        \includegraphics[width=12cm]{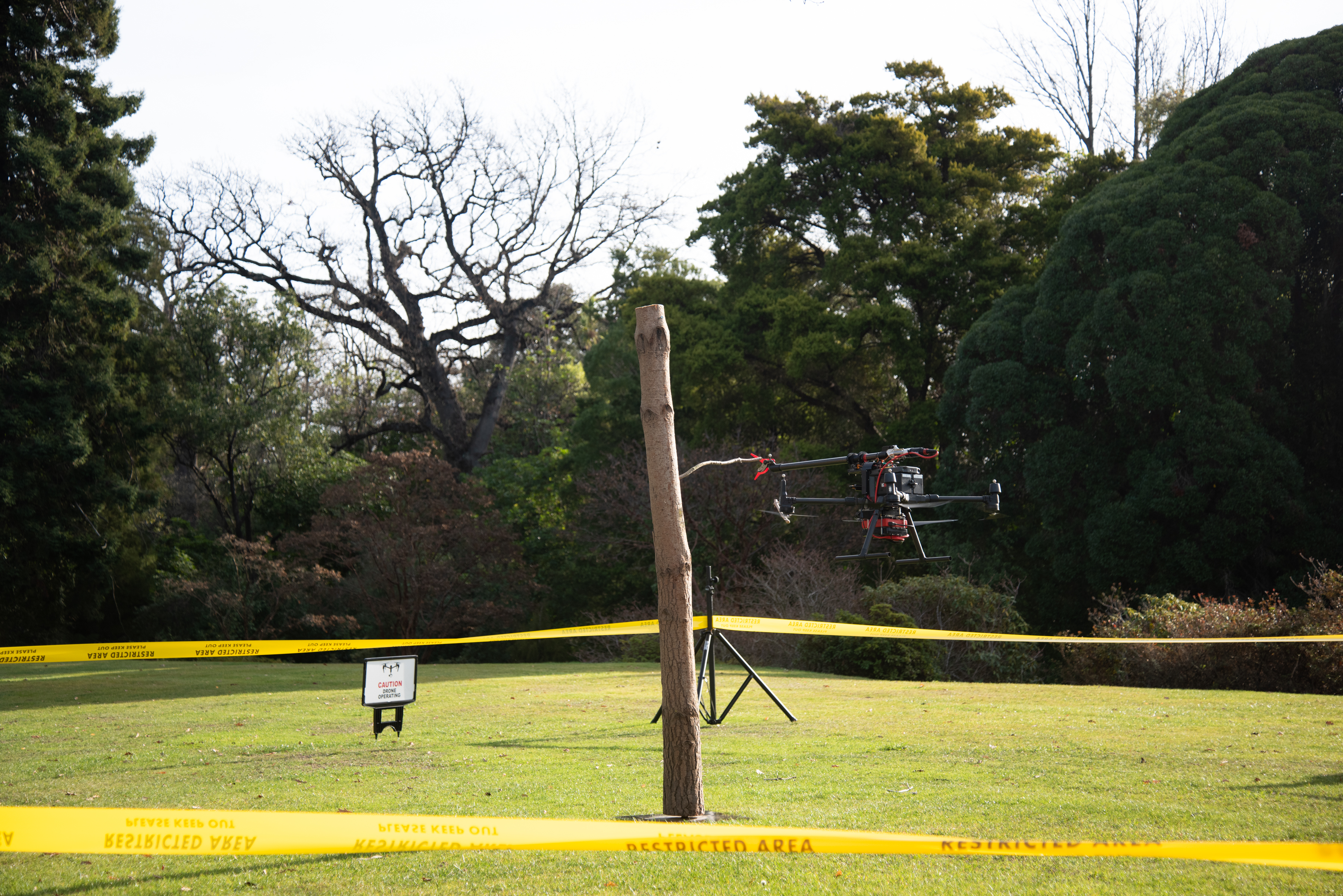}
        \caption{The drone, equipped with a ZED Mini camera for stereo vision and
        a robotic arm, autonomously detects and prunes branches of radiata pine.
        The ZED Mini camera enables the drone to accurately identify the branches,
        while the robotic arm precisely prunes them.}
        \label{drone}
    \end{figure}

    For a detailed overview of the drone hardware used in this study, see \url{https://ucvision.org.nz/drones/};
    the design and testing process are depicted in Figure~\ref{drone}.

    This multi-institutional project brings together several research groups, each
    contributing distinct expertise. While other institutions focus on the
    physical construction of the drones, our research is concentrated on developing
    the vision detection and measurement algorithms for the cameras mounted on these
    drones. These algorithms detect branches and determine their three-dimensional
    positions, information that is essential for guiding the drone's robotic arm
    during pruning. The primary objective is to improve the accuracy and
    reliability of branch detection and positioning, thereby increasing the efficiency
    and safety of autonomous pruning.

    \section{Related Work}


    In contemporary agricultural machinery, pruning equipment is often associated
    with high costs and the integration of advanced technologies, such as
    multiple high-precision cameras and LiDAR sensors \citep{kulbacki2018survey,pakeerathan2023smart}.
    While these systems significantly enhance operational efficiency, they also contribute
    to increased costs and complexity in deployment, which may impede the
    commercial viability and broader adoption of drone-based solutions. To mitigate
    these challenges, this research emphasizes the use of a single stereo camera
    for branch detection and distance measurement between branches and the drone.

    Given the reliance on a single stereo camera, the study leverages advanced computer
    vision techniques to detect the branches, with a particular focus on two
    critical areas: branch detection\citep{zou2023object} and depth map generation\citep{laga2020survey}.
    For branch detection, the algorithms must be both highly efficient and precise,
    as the drone needs to identify branches in real-time during operation.

    To facilitate depth map generation, innovative methods are employed to accurately
    estimate the distance between the branches and the camera. By integrating
    branch detection with depth map generation, the spatial positioning of branches
    can be precisely determined. This accurate spatial information then enables
    the drone's robotic arm to execute branch trimming with precision, ensuring effective
    and targeted pruning.

    \subsection{Detection and Segmentation}

    Object detection \citep{zou2023object} and image segmentation \citep{minaee2021image}
    are both critical tasks in computer vision. Object detection primarily focuses
    on identifying and locating objects within an image, typically by marking
    their positions with bounding boxes. Segmentation, on the other hand, takes this
    a step further by dividing the image into distinct regions, accurately
    delineating the shape and boundaries of objects.

    In this study, the focus extends beyond merely identifying the positional information
    of tree branches to include the acquisition of detailed locational data of surrounding
    points. This requirement necessitates a transition from conventional object
    detection methods to more precise image segmentation techniques \citep{sharma2022survey}.
    By utilizing segmentation on drone-captured imagery, this research seeks to accurately
    ascertain the precise location of tree branches and their neighboring regions.

    The evolution of object detection and image segmentation has been marked by significant
    advancements since the introduction of the Region-based Convolutional Neural
    Network (R-CNN) in 2014 \citep{girshick2014rich}. R-CNN represented a leap forward
    in detection accuracy by utilizing candidate regions for feature extraction and
    classification. The Spatial Pyramid Pooling Net (SPP-Net) \citep{he2015spatial}
    addressed the fixed input size constraint, allowing networks to retain more
    spatial information and improving the efficiency of feature extraction.

    Fast R-CNN \citep{girshick2015fast} improved on this by integrating the ROI Pooling
    layer, which enabled feature extraction directly on the feature map and
    significantly shortened training time. Faster R-CNN, through its Region
    Proposal Network (RPN) \citep{ren2016faster}, allowed candidate-region
    generation and feature extraction to share computational resources, boosting
    both speed and accuracy. Mask R-CNN \citep{he2017mask} added an additional branch
    for generating object masks, enabling pixel-level segmentation.

    For real-time object detection, the YOLO series \citep{redmon2016you, reis2023real, wang2024yolov9}
    has established itself as a highly influential framework within both
    industrial and academic contexts, primarily due to its remarkable speed and precision.
    Considering its applicability to drone-based operations, the evaluation of the
    latest YOLO algorithm is prioritised in the experiments that follow.

    In evaluating the YOLO segmentation model, it is crucial to consider two
    essential sub-indicators: computational efficiency, particularly the running
    time, as the drone must execute image interception and segmentation with exceptional
    speed; and accuracy, which we evaluate using mAP 50-95 (Mean Average
    Precision at IoU thresholds ranging from 50\% to 95\%). We define $AP(t)$ as
    the Average Precision at a specific IoU threshold $t$, where $t$ represents the
    IoU threshold varying from 0.5 to 0.95 in increments of 0.05. The variable
    $n$ denotes the total number of IoU thresholds, which equals 10 as specified
    by the COCO evaluation protocol\citep{lin2014microsoft}. The formula for mAP$^{\text{50-95}}$
    is given by:
    \begin{equation}
        \text{mAP}_{50-95}= \frac{1}{n}\sum_{t=0.5}^{0.95}\text{AP}(t)
    \end{equation}

    \subsection{Depth Map}
    Depth map generation \citep{laga2020survey} is a critical component of
    computer vision, involving the inference of a scene's three-dimensional structure
    from one or more images. In applications such as precise trimming with a
    robotic arm on a drone, accurate determination of both the spatial location of
    the tree branch and its distance from the drone is essential.

    Depth maps, representing the distance from each pixel in the image to the
    camera, are generated using either active or passive methods. Active methods
    employ sensors that emit and receive signals to measure depth, including technologies
    such as LiDAR\citep{wu2018squeezeseg}, structured light\citep{yang2024overview},
    and time-of-flight cameras\citep{kolb2010time}. Passive methods, by contrast,
    rely on existing optical information through techniques such as stereo
    matching\citep{wang2011obtaining}, multi-view geometry\citep{hartley2003multiple},
    and monocular depth estimation\citep{eigen2014depth}.

    Since the drone is equipped exclusively with a stereo camera, the methodology
    is inherently constrained to the use of two or fewer cameras. Consequently, stereo
    matching generates depth maps by deriving values through triangulation,
    leveraging the parallax effect between the cameras. The relevant formulas are
    detailed in the subsequent sections.

    \begin{figure}[htbp]
        \centering
        \includegraphics[width=12cm]{1.png}
        \caption{Triangulation using Two Cameras to Obtain the Depth Map. The point
        $(u_{l}, v_{l})$ represents the projection of point $p (x, y, z)$ in three-dimensional
        space onto the image plane of the left camera, whereas point
        $(u_{r}, v_{r})$ corresponds to the projection of the same point onto
        the right camera’s image plane. The variable $b$ denotes the baseline distance
        separating the left and right cameras.}
        \label{fig:triangulation1}
    \end{figure}

    \begin{figure}[htbp]
        \centering
        \includegraphics[width=12cm]{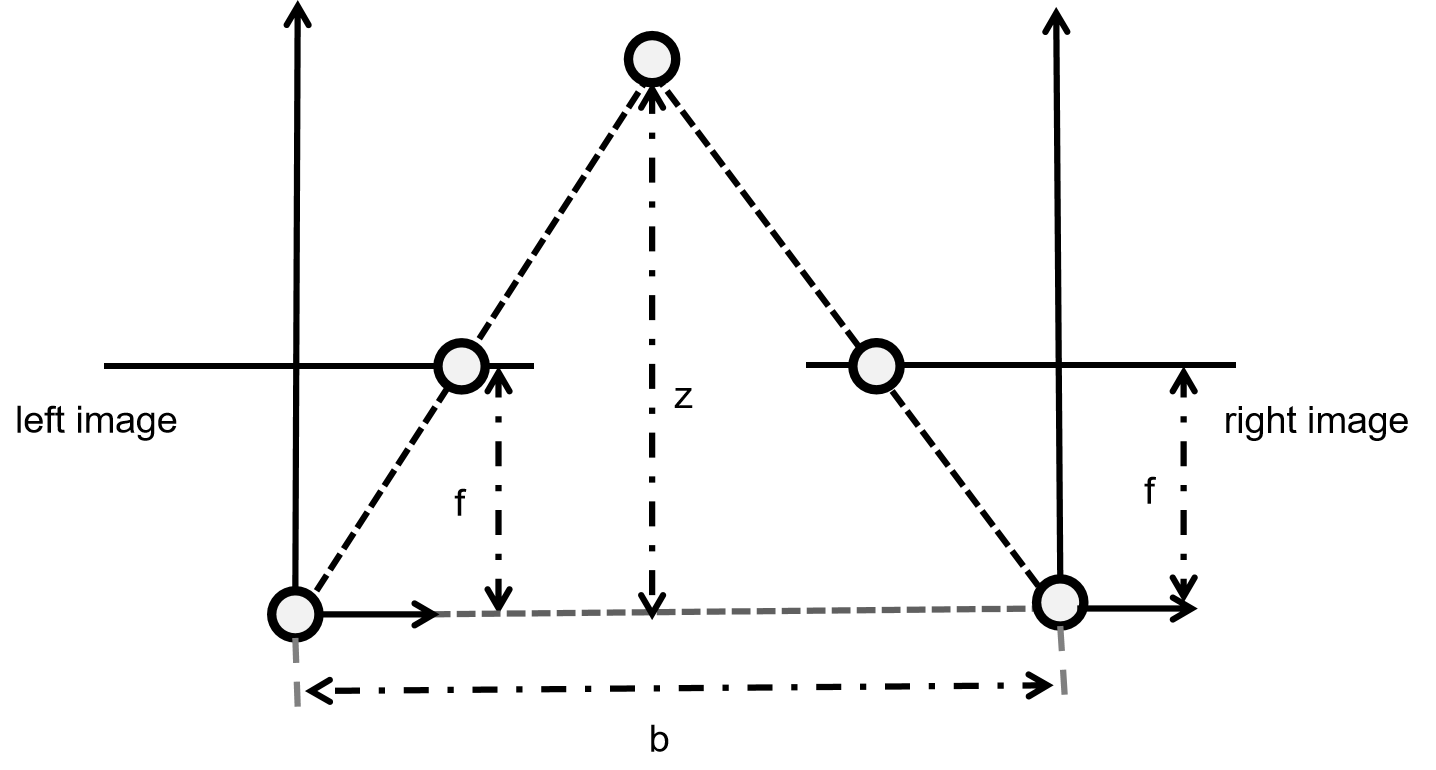}
        \caption{Triangulation using Two Cameras to Obtain the Depth Map. $f$ represents
        the camera's focal length. With the values of $f$, $b$ (the baseline
        distance between the cameras), and the disparity of point $p (x, y, z)$
        between the left and right camera images, we can calculate the distance $z$
        from the pixel representing point $p$ in space to the camera.}
        \label{fig:triangulation2}
    \end{figure}

    \noindent
    $f_{x}$: the focal length of the camera in the $x$-direction.\\
    $f_{y}$: the focal length of the camera in the $y$-direction. \\
    $o_{x}$: the horizontal offset of the principal point (the projection of the
    optical centre) from the top-left corner of the image sensor, i.e.~the horizontal
    distance from the sensor's left edge to the image centre.\\
    $o_{y}$: the vertical offset of the principal point from the top-left corner
    of the image sensor, i.e.~the vertical distance from the sensor's top edge
    to the image centre.\\
    $u_{l}$: the horizontal coordinate of a point in the left image. \\
    $u_{r}$: the horizontal coordinate of the corresponding point in the right image
    for the same scene point as in the left image. \\
    $v_{l}$: the vertical coordinate of a point in the left image.\\
    $v_{r}$: the vertical coordinate of the corresponding point in the right
    image for the same scene point as in the left image. \\
    $b$: the baseline distance between the centers of two cameras in a stereo
    setup.\\
    $d$: the disparity, which is the horizontal position difference of the same
    scene point between the left and right images.\\
    $z$: the depth value, representing the distance from the object point to the
    camera.

    Definitional equations (\ref{eq:ul}) and (\ref{eq:vl}) describe the position
    of pixel point $(u_{l}, v_{l})$ in the left camera view, while equations (\ref{eq:ur})
    and (\ref{eq:vr}) describe the position of pixel point $(u_{r}, v_{r})$ in the
    right camera view.

    \begin{equation}
        u_{l}= f_{x}\frac{x}{z}+o_{x}\label{eq:ul}
    \end{equation}

    \begin{equation}
        v_{l}= f_{y}\frac{y}{z}+o_{y}\label{eq:vl}
    \end{equation}

    \begin{equation}
        u_{r}= f_{x}\frac{x-b}{z}+o_{x}\label{eq:ur}
    \end{equation}

    \begin{equation}
        v_{r}= f_{y}\frac{y}{z}+o_{y}\label{eq:vr}
    \end{equation}

    \noindent
    Combining Eqs.~(\ref{eq:ul}) and (\ref{eq:vl}), as well as Eqs.~(\ref{eq:ur})
    and (\ref{eq:vr}), we obtain the pixel coordinates as Eq.~(\ref{eq:combined}).
    \begin{equation}
        \begin{aligned}
            (u_{l},v_{l}) = (f_{x}\frac{x}{z}+o_{x}, f_{y}\frac{y}{z}+o_{y})   \\
            (u_{r},v_{r}) = (f_{x}\frac{x-b}{z}+o_{x}, f_{y}\frac{y}{z}+o_{y})
        \end{aligned}
        \label{eq:combined}
    \end{equation}
    \noindent
    Based on the pixel coordinates of the left and right cameras, we can find
    the coordinates of the object in three dimensions $(x,y,z)$.
    \begin{equation}
        \begin{aligned}
            x = \frac{b(u_{l}-o_{x})}{(u_{l}-u_{r})}           \\
            y = \frac{bf_{x}(v_{l}-o_{y})}{f_{y}(u_{l}-u_{r})} \\
            z = \frac{bf_{x}}{(u_{l}-u_{r})}
        \end{aligned}
    \end{equation}

    \noindent
    From the above, the disparity $d$ and depth $z$ are defined as:

    \begin{equation}
        d = u_{l}- u_{r}
    \end{equation}

    \begin{equation}
        z = \frac{b f_{x}}{u_{l}-u_{r}}\label{eq:depth}
    \end{equation}

    \noindent
    We set the product of the baseline $b$ and the focal length of the camera in
    the x-direction $f_{x}$ to a new variable $W$.
    \begin{equation}
        W = b \cdot f_{x}
    \end{equation}
    Substituting $W$ into Eq.~(\ref{eq:depth}), we get the more concise Eq.~(\ref{eq:depth_concise}).
    \begin{equation}
        z = \frac{W}{d}\label{eq:depth_concise}
    \end{equation}
    Since $W$ is fixed, and $z$ and $d$ are inversely proportional, the larger
    the disparity value, the smaller the depth value. In other words, a larger disparity
    value indicates that the pixel point is closer to the camera. Among
    traditional methods, Block Matching (BM) \citep{scharstein2002taxonomy} and
    Semi-Global Block Matching (SGBM) \citep{hirschmuller2007stereo} are two dominant
    techniques. BM is a local search-based method that calculates depth values
    by finding the best match within a fixed window, making it suitable for real-time
    applications, though it is prone to errors in sparse texture or overlapping
    regions. In contrast, the SGBM method introduces a semi-global cost
    aggregation strategy, improving the accuracy and robustness of depth estimation
    by optimizing pixel points across the entire image, particularly effective
    in handling texture-rich scenes.

    With the development of deep learning techniques, neural network-based depth
    map generation methods have significantly improved the accuracy and efficiency
    of depth estimation. For example, ACVNet \citep{xu2022attention} and GWCNet
    \citep{guo2019group} introduce more sophisticated learning mechanisms to process
    stereo images, and generate more accurate depth maps by learning matching
    information from paired images through deep learning models. MobileStereoNet
    \citep{shamsafar2022mobilestereonet} optimises the network architecture for
    mobile devices and achieves efficient depth prediction. PSMNet \citep{chang2018pyramid}
    enhances feature extraction and depth estimation through pyramid pooling and
    3D convolution, while RAFT-Stereo \citep{lipson2021raft} employs iterative
    updates via gated recurrent units to perform pixel-level correlation
    refinement, further improving the accuracy of depth estimation. NeRF-Supervised
    Deep Stereo \citep{tosi2023nerf} demonstrates an innovative approach that trains
    stereo matching networks using synthetic data rendered from Neural Radiance Fields,
    which is particularly useful for scenes where ground-truth depth annotations
    are unavailable. Through the application of traditional methods and deep neural
    network architectures, depth maps for tree branches are generated. The quality
    of these depth maps is evaluated using metrics such as Root Mean Squared Error.

    \begin{equation}
        \text{RMSE}= \sqrt{\frac{1}{n}\sum_{i=1}^{n}(y_{i}- \hat{y}_{i})^{2}}
    \end{equation}

    where:
    \begin{itemize}
        \item $n$ is the total number of data points.

        \item $y_{i}$ represents the actual value.

        \item $\hat{y}_{i}$ represents the predicted value.
    \end{itemize}

    The evaluation is performed on various public datasets, followed by fine-tuning
    on the specific tree branch dataset. A detailed analysis is presented in
    Section 4.

    \section{Methods}

    \subsection{Data Collection and Image Segmentation}
    Currently, there are many efficient algorithms on the market that can accurately
    perform image segmentation, so the training and testing phase is no longer
    the focus of the study. In this study, we mainly collected indoor data by
    using ZED Mini \url{https://store.stereolabs.com/products/zed-mini} cameras
    in different corners of the laboratory, with a resolution set to 1920 x 1080.
    We took photos of various tree branches under different lighting conditions to
    avoid over-idealisation of the training images. So far, 61 pairs of photos (i.e.
    122 photos) have been collected for the training dataset, and another 10 pairs
    of photos have been collected for the test dataset.

    After the data has been collected, our main task is to annotate these images
    to identify key points around the branch that should completely surround the
    branch. After labelling, we input the data into different segmentation
    models for processing. Even with a model with small parameters, we were able
    to obtain satisfactory results. This is mainly due to the small size and simple
    structure of our dataset, which contains only a few basic branches, thus making
    the processing more efficient.

    \begin{table}[h!]
        \centering
        \begin{tabular}{lccccc}
            \toprule \textbf{model}    & \multicolumn{2}{c}{\textbf{COCO datasets}} & \multicolumn{2}{c}{\textbf{Branches}} \\
            \midrule                   & mAP\textsubscript{box50--95}               & mAP\textsubscript{mask50--95}        & mAP\textsubscript{box50--95} & mAP\textsubscript{mask50--95} \\
            \midrule Mask R-CNN R50-C4 & 39.8                                       & 34.4                                 & 76.86                        & 0.06                          \\
            Mask R-CNN R50-DC5         & 40.0                                       & 35.9                                 & 77.54                        & 9.16                          \\
            Mask R-CNN R50-FPN         & 41.0                                       & 37.2                                 & 79.19                        & 6.75                          \\
            Mask R-CNN R101-C4         & 42.6                                       & 36.7                                 & 88.05                        & 0.05                          \\
            Mask R-CNN R101-DC5        & 41.9                                       & 37.3                                 & 79.12                        & 9.94                          \\
            Mask R-CNN R101-FPN        & 42.9                                       & 38.6                                 & 84.09                        & 2.95                          \\
            Mask R-CNN X101-FPN        & 44.3                                       & 39.5                                 & 85.52                        & 11.55                         \\
            YOLOv8n-seg                & 36.7                                       & 30.5                                 & 98.9                         & 77.4                          \\
            YOLOv8s-seg                & 44.6                                       & 36.8                                 & 99.5                         & 82.0                          \\
            YOLOv8m-seg                & 49.9                                       & 40.8                                 & 99.6                         & 81.6                          \\
            YOLOv8l-seg                & 52.3                                       & 42.6                                 & 99.2                         & 80.1                          \\
            YOLOv8x-seg                & 53.4                                       & 43.4                                 & 98.7                         & 77.1                          \\
            YOLOv9c-seg                & 52.4                                       & 42.2                                 & 98.9                         & 80.9                          \\
            YOLOv9e-seg                & 55.1                                       & 44.3                                 & 98.8                         & 80.0                          \\
            \bottomrule
        \end{tabular}
        \caption{Performance Comparison of Mask R-CNN with Different Backbones
        and Various YOLO Segmentation Models. For Mask R-CNN, the FPN backbone, using
        ResNet with conv and FC heads, offers the best speed-accuracy tradeoff. The
        C4 backbone, based on ResNet conv4 with a conv5 head, follows the original
        Faster R-CNN design, while the DC5 variant, with dilations in ResNet conv5,
        is inspired by Deformable ConvNets. The mAPbox and mAPmask metrics represent
        the model’s average accuracy in detection and segmentation tasks across IoU
        thresholds from 50\% to 95\%.}
        \label{tab:yolo}
    \end{table}

    The experimental results on the COCO dataset demonstrate that the majority
    of models exhibit strong performance in both object detection and image
    segmentation. Specifically, when evaluated using the mAPbox and mAPmask
    metrics, models such as Mask R-CNN and the YOLO series achieve high levels of
    accuracy in detecting objects and performing segmentation. Due to the
    simplicity of the dataset, the task of object detection was almost entirely resolved,
    which may have limited the ability of the models to fully showcase their
    capabilities. It is also important to note that each model was trained for only
    100 epochs, primarily due to the small size of the dataset.

    A more detailed analysis of the results reveals that Mask R-CNN models
    maintain stable performance across different configurations, particularly in
    the mAP\textsubscript{box50--95} and mAP\textsubscript{mask50--95} metrics,
    with only slight variations observed between the various model architectures.
    In contrast, the YOLOv8 series stands out in both detection and segmentation
    tasks, achieving near-perfect performance in the mAP\textsubscript{box50--95}
    metric, with values such as 98.9 for YOLOv8n-seg and over 98.7 for YOLOv8x-seg,
    highlighting their remarkable detection accuracy. However, given the limited
    complexity and size of the dataset, these models’ true performance in more
    complex scenarios may not have been fully revealed.

    These findings underscore the notable strengths of the YOLO series in both
    object detection and segmentation tasks, even with relatively few training iterations.
    However, as more data is collected from real-world environments,
    particularly through the use of drones, the limitations of these models,
    especially when addressing more complex objects and environments, are likely
    to become more pronounced. This emphasizes the need for an expanded dataset
    to comprehensively assess model performance in more challenging contexts.

    \subsection{Traditional Methods for Generating a Disparity Map}

    As established in Eq.~(\ref{eq:depth_concise}), producing an accurate depth map
    hinges on the quality of the underlying disparity map --- if the disparity map
    is sufficiently accurate, the depth map will be too, since the baseline distance
    $b$ and focal length $f_{x}$ are fixed camera parameters. The traditional
    approach to generating a disparity map involves four stages. The relevant parameters
    and the BM (Block Matching) algorithm are defined below. $E_{l}$: The left camera
    image\\
    $E_{r}$: The right camera image\\
    $L$: the template window, a small region in the left camera image $E_{l}$
    used to search for matches in $E_{r}$.\\
    $T$: the search scan line, the horizontal line along which $E_{r}$ is
    scanned to find the region best matching the template window $L$.\\

    In BM (Block Matching), a specific area in one image is selected as a
    template, and the most similar area is searched for in another image. By
    comparing the similarity of pixels within these two areas, such as calculating
    the differences in their pixel intensities, we can determine the position of
    the best match. The positional difference between the matched areas, known
    as disparity, is then used to estimate the depth of various objects in the scene.

    \begin{figure}[htbp]
        \centering
        \includegraphics[width=14cm]{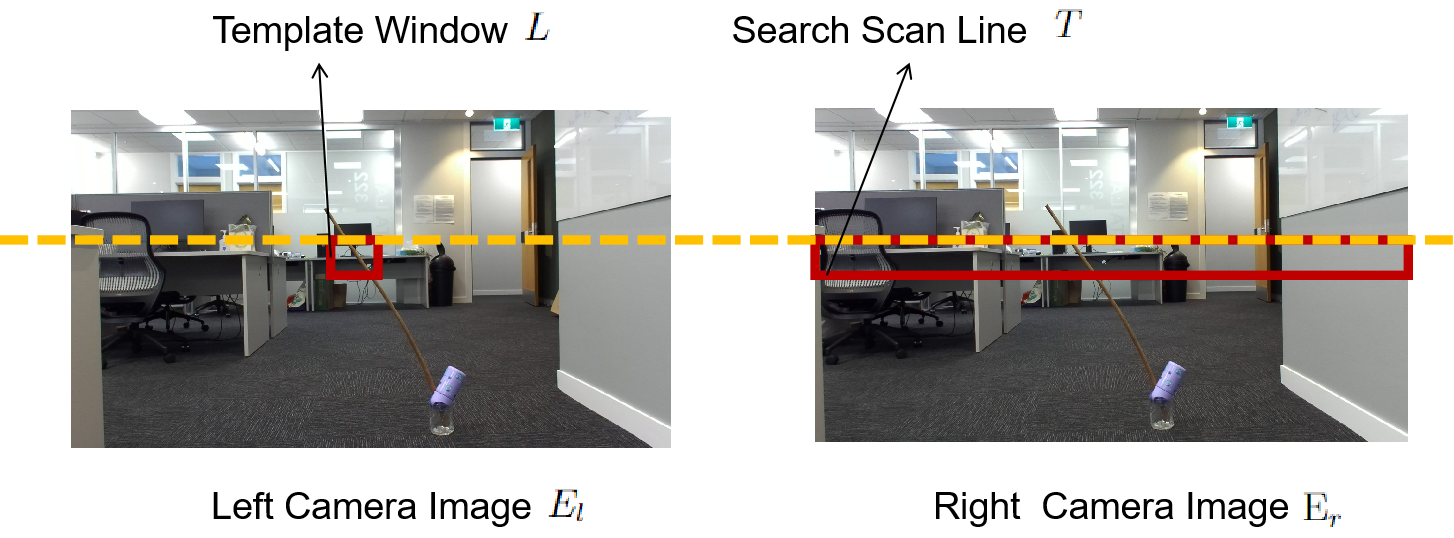}
        \caption{Block Matching illustration. $L$ denotes the template window and
        $T$ the search scan line. The left camera image $E_{l}$ serves as the
        reference, and the corresponding pixel is located in $E_{r}$.}
        \label{fig:block_matching}
    \end{figure}
    Matching cost calculation determines how well each pixel in the left image corresponds
    to a candidate pixel in the right image. Common formulations include AD (Absolute
    Intensity Differences), SD (Squared Intensity Differences), and NCC (Normalized
    Cross-Correlation). Here we assume that $E_{l}(x,y)$ represents the pixel intensity
    value at position $(x, y)$ in the left image, and $E_{r}(x+d,y)$ represents
    the pixel intensity value at position $(x+d, y)$ in the right image, where $d$
    is the disparity. These formulas are as follows.

    AD calculates the absolute intensity difference between corresponding pixel
    points in the left and right images.
    \begin{equation}
        AD(x, y, d) = |E_{l}(x, y) - E_{r}(x + d, y)|
    \end{equation}

    SD calculates the squared intensity difference between corresponding pixel
    points in the left and right images, and squaring the differences makes them
    positive and penalizes larger differences more heavily.
    \begin{equation}
        SD(x, y, d) = \left( E_{l}(x, y) - E_{r}(x + d, y) \right)^{2}
    \end{equation}

    NCC is a method to calculate the match between two images in a local area,
    which is more robust against brightness changes.
    \begin{equation}
        \begin{split}
            NCC(x, y, d) = \frac{\sum_{(i,j) \in T}\left( E_{l}(i, j) - \mu_{L}\right)
            \left( E_{r}(i + d, j) - \mu_{R}\right)}{\sqrt{\sum_{(i,j) \in T}\left(
            E_{l}(i, j) - \mu_{L}\right)^{2}\sum_{(i,j) \in T}\left( E_{r}(i + d,
            j) - \mu_{R}\right)^{2}}}\\
        \end{split}
    \end{equation}
    $\mu_{L}$ and $\mu_{R}$ denote the mean intensity values of the left and right
    image patches within the window $T$, respectively.
    \begin{equation}
        \begin{split}
            \mu_{L}= \frac{1}{|T|}\sum_{(i,j) \in T}E_{l}(i, j) \\
            \mu_{R}= \frac{1}{|T|}\sum_{(i,j) \in T}E_{r}(i + d, j)
        \end{split}
    \end{equation}

    Cost aggregation combines values from multiple directions, using a penalty parameter
    to smooth the results and balance local and global information.

    Fixed Window aggregation sums or averages the matching cost within a fixed-size
    window.

    \begin{equation}
        C_{fixed}(x, y, d) = \sum_{(i,j) \in T}C(i, j, d)
    \end{equation}

    Multiple Windows aggregation uses multiple windows at different locations, where
    $T_{k}$ is the $k^{th}$ window and there are $N$ windows in total.
    \begin{equation}
        C_{multi}(x, y, d) = \sum_{k=1}^{N}\sum_{(i,j) \in T_k}C(i, j, d)
    \end{equation}

    Iterative Diffusion propagates matching costs across the neighbourhood over multiple
    iterations. $w_{n}$ is the weight, and $C_{diff}^{(n-1)}$ is the cost from the
    $(n-1)^{th}$ iteration.

    \begin{equation}
        C_{diff}(x, y, d) = \sum_{n=1}^{N}w_{n}\cdot C_{diff}^{(n-1)}(x, y, d)
    \end{equation}

    Minimum cost selection sums the aggregated costs across all directions and
    assigns to each pixel the disparity value with the smallest total cost, yielding
    the most reliable disparity map.

    Local Methods:
    \begin{equation}
        d(x, y) = \arg \min_{d}C(x, y, d)
    \end{equation}

    Global Methods:
    \begin{equation}
        E(d) = E_{data}(d) + \lambda E_{smooth}(d)
    \end{equation}

    \begin{equation}
        E_{data}(d) = \sum_{(x,y)}C(x, y, d(x, y))
    \end{equation}
    \begin{equation}
        \begin{split}
            E_{smooth}(d) = \sum_{(x,y)}\rho(d(x, y) - d(x+1, y)) + \rho(d(x, y)
            - d(x, y+1))
        \end{split}
    \end{equation}

    Post-processing completes the pipeline through left-right consistency
    checking, disparity map filtering, and sub-pixel accuracy enhancement.
    \begin{equation}
        \begin{split}
            d_{sub}(x, y) = d(x, y) - \frac{C(x, y, d+1) - C(x, y, d-1)}{2 \left(
            C(x, y, d+1) + C(x, y, d-1) - 2C(x, y, d) \right)}
        \end{split}
    \end{equation}
    \begin{equation}
        d_{post}(x, y) = \text{median}\{ d(x+i, y+j) | (i, j) \in W \}
    \end{equation}

    \noindent
    In addition to the above post-processing methods, we can also use weighted least
    squares (WLS) to optimise the disparity results, and here we use the left
    image as an example and define the following parameters:

    \noindent
    $d'$: The optimized disparity map.\\
    \noindent
    $d(x, y)$: The initial disparity map.\\
    \noindent
    $w_{(x,y)}$: The weight of pixel $(x, y)$.\\
    \noindent
    $\lambda$: The regularization parameter. This controls the balance between the
    data term and the smoothness term. A larger $\lambda$ value makes the
    smoothness term more influential, resulting in a smoother disparity map.\\
    \noindent
    $w_{(x,y),(x',y')}$: The weight between pixel $(x, y)$ and pixel $(x', y')$.\\
    \noindent
    $\sigma$: The smoothness parameter. This controls the degree of smoothing in
    the color space. A smaller $\sigma$ value makes the filter more sensitive to
    color changes, helping to preserve edges and details.\\

    The optimization formula for the disparity map $d$ is given by:
    \begin{equation}
        \begin{split}
            d' = \arg \min_{d}\Bigg(\sum_{(x,y)}w_{(x,y)}(d'(x,y) - d(x,y))^{2}+
            \lambda\sum_{(x,y),(x',y')}w_{(x,y),(x',y')}(d'(x,y) - d'(x',y'))^{2}
            \Bigg)
        \end{split}
    \end{equation}

    where the weight calculation formula is:
    \begin{equation}
        w_{(x,y),(x',y')}= \exp \left( - \frac{(E_{l}(x,y) - E_{l}(x',y'))^{2}}{2\sigma^{2}}
        \right)
    \end{equation}

    The left-right consistency check detects and removes inconsistent matches,
    disparity filtering smoothes the disparity map to reduce noise, and sub-pixel
    accuracy enhancement further improves the accuracy of the disparity map through
    interpolation. Here we use SGBM as a case study to show the steps of the
    traditional method to generate the disparity map, and the result is shown in
    the following figure:

    \begin{figure}[htbp]
        \centering
        \subfigure[original left image]{\includegraphics[width=0.3\textwidth]{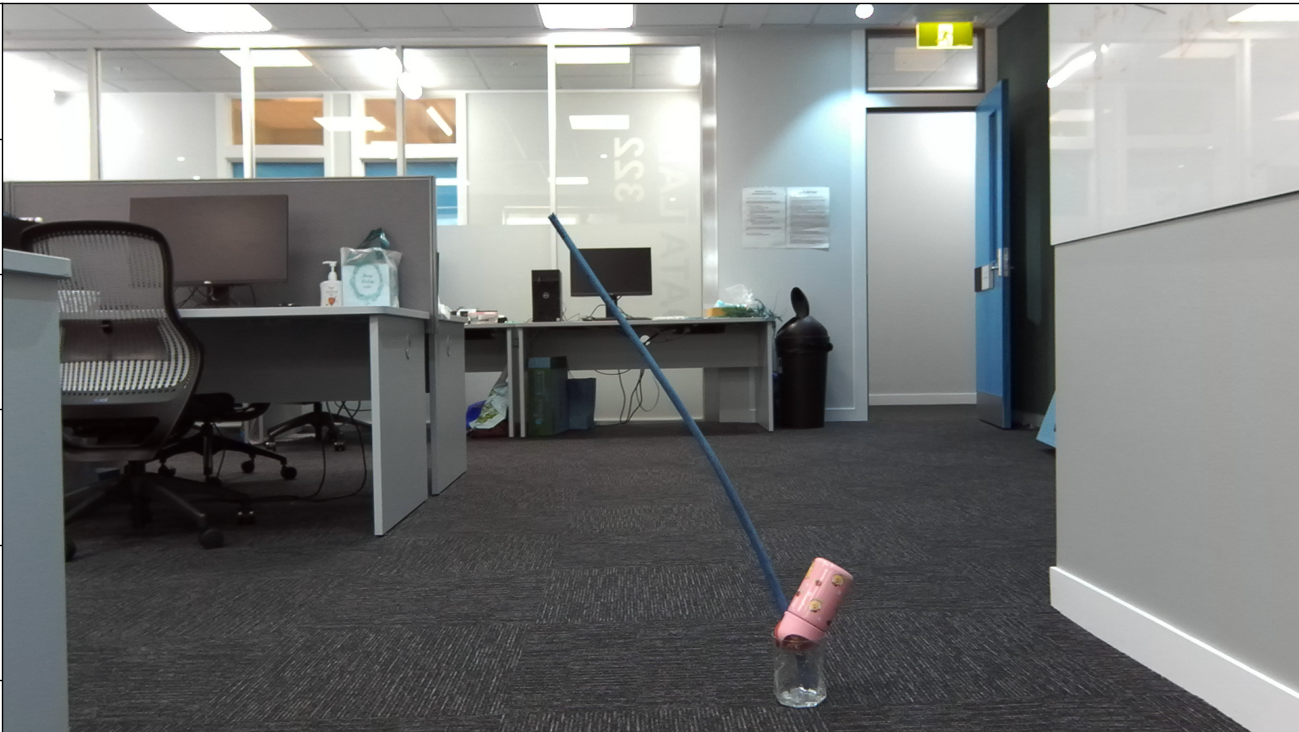}}
        \subfigure[original right image]{\includegraphics[width=0.3\textwidth]{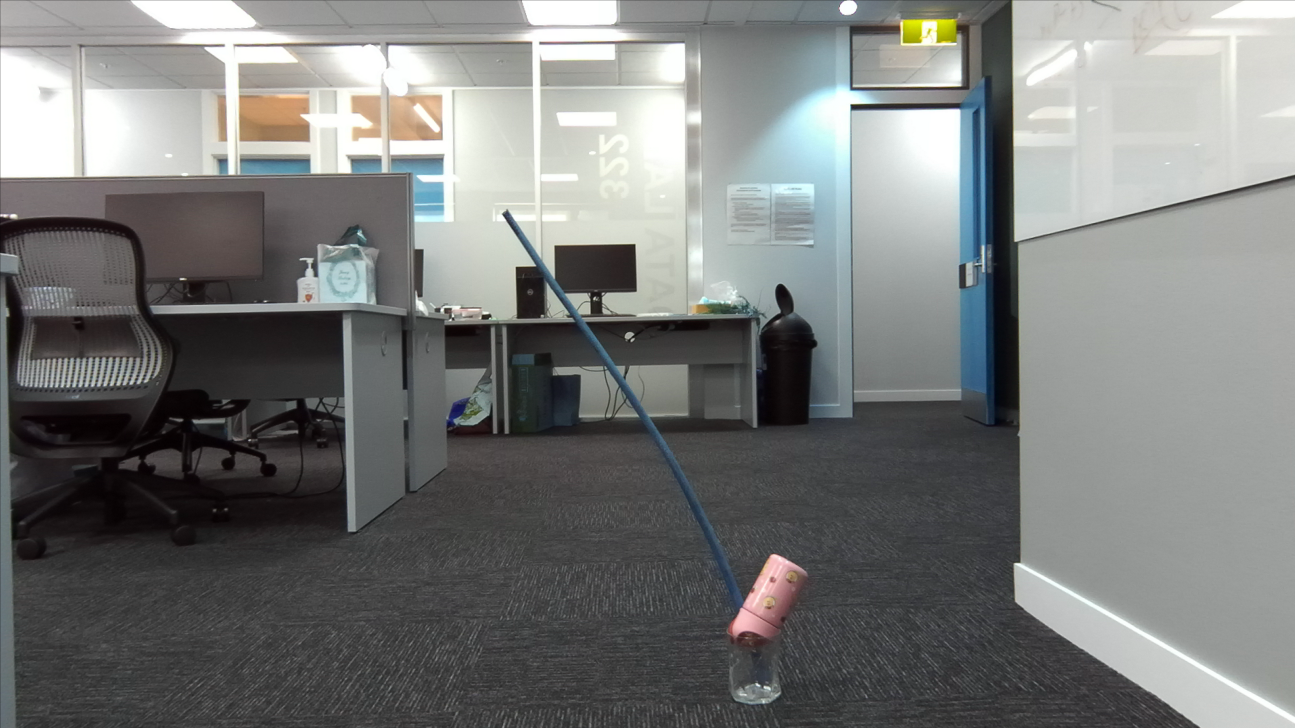}}
        \subfigure[left image after pre-processing]{\includegraphics[width=0.3\textwidth]{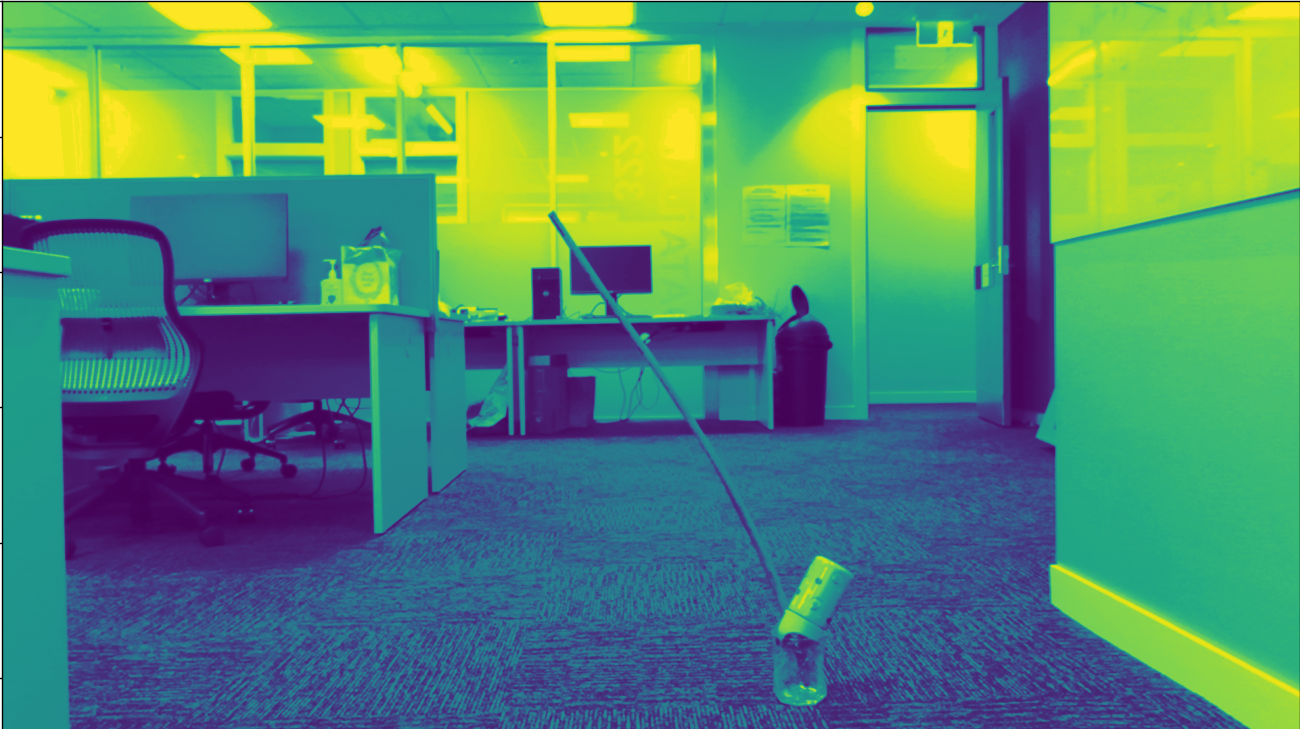}}
        \subfigure[right image after pre-processing]{\includegraphics[width=0.3\textwidth]{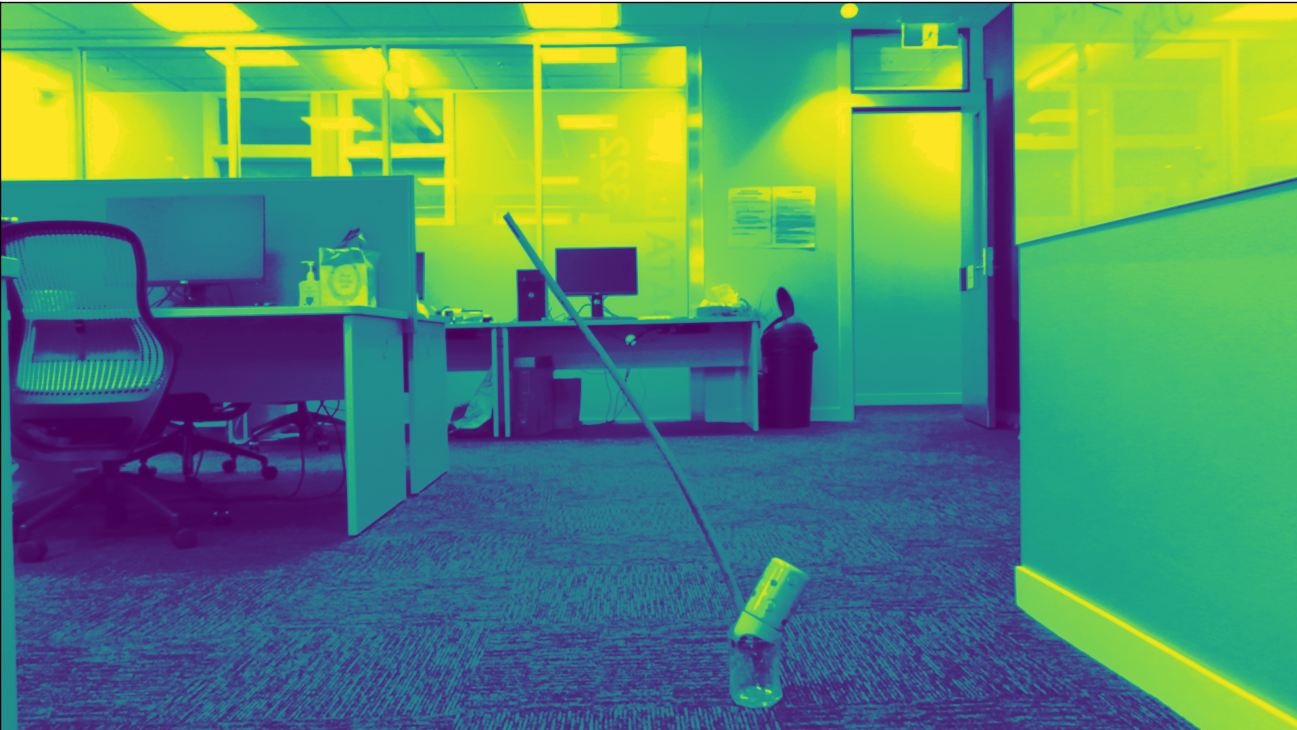}}
        \subfigure[disparity map through SGBM]{\includegraphics[width=0.3\textwidth]{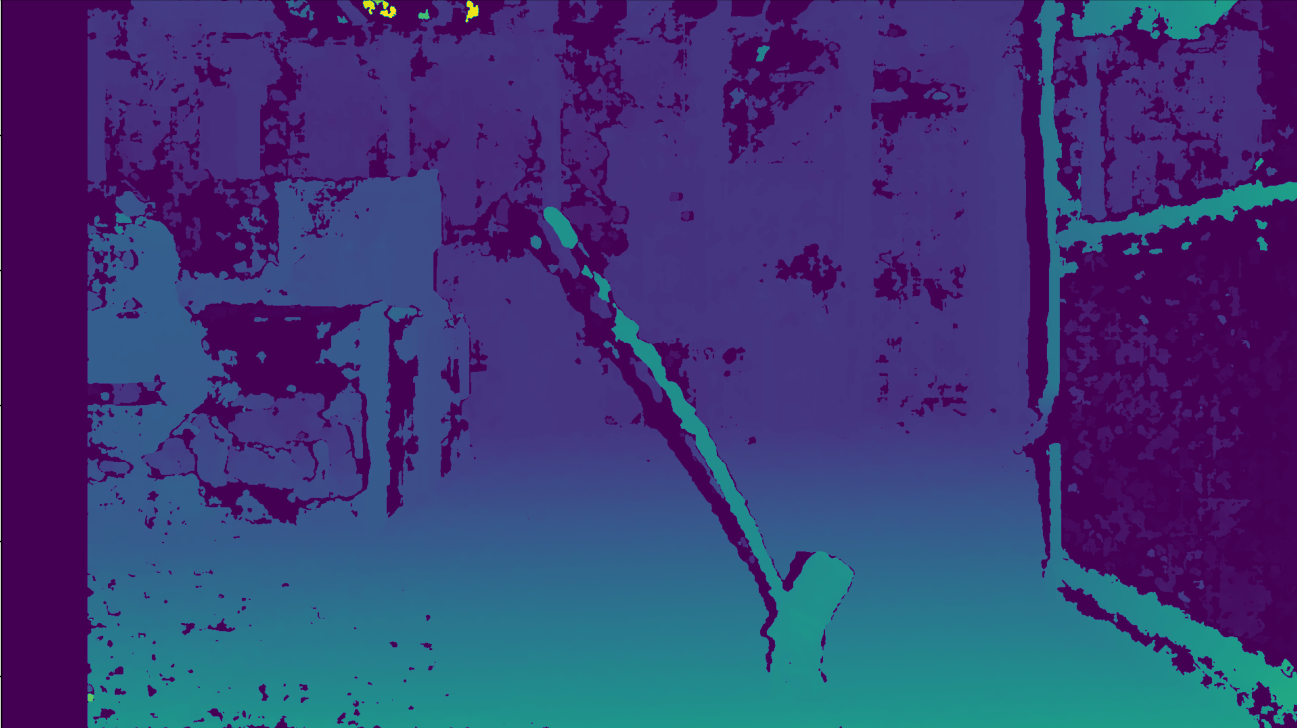}}
        \subfigure[disparity map processed by Weighted Least Squares]{\includegraphics[width=0.3\textwidth]{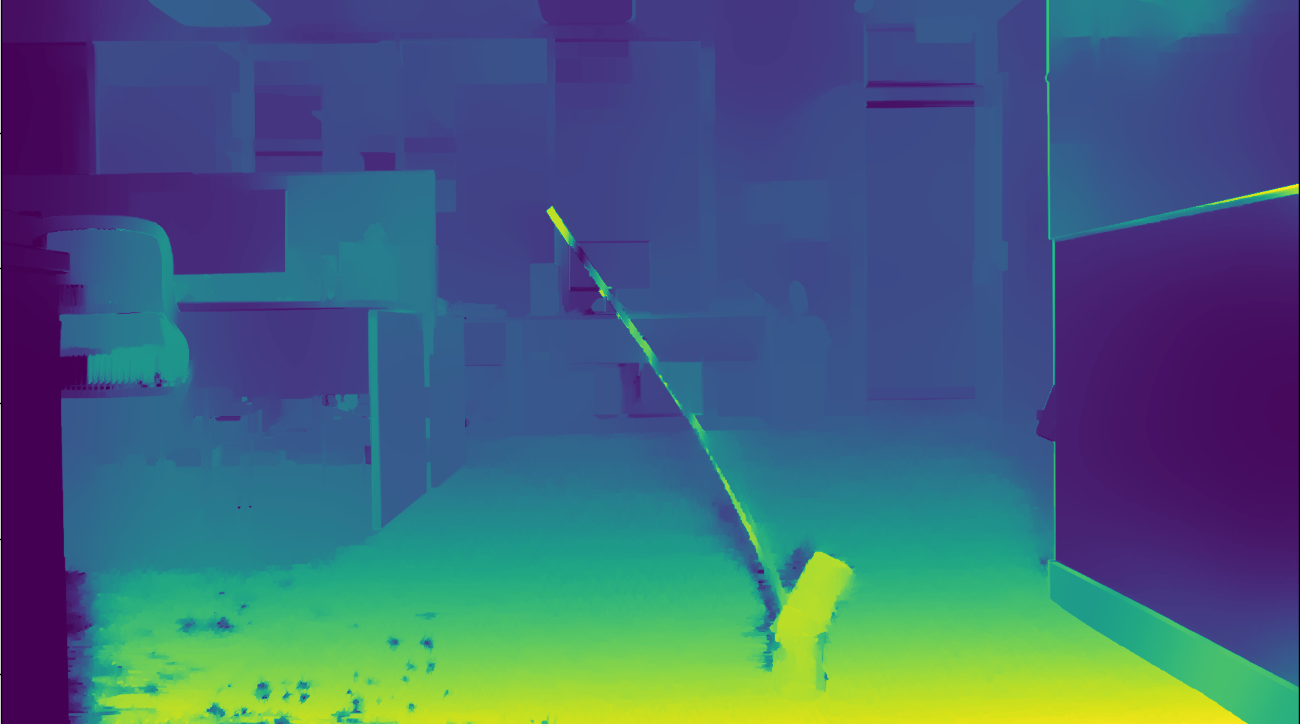}}
        \caption{SGBM disparity map generation pipeline: (a,\,b) original left
        and right images, (c,\,d) pre-processed images, (e) SGBM-generated disparity
        map, and (f) WLS-refined disparity map.}
        \label{fig:sgbm_pipeline}
    \end{figure}

    \subsection{Deep Learning Methods for Generating a Disparity Map}

    Deep learning-based disparity map generation can be divided into monocular and
    multi-view approaches. This work focuses on binocular methods, where a network
    predicts per-pixel disparity from a rectified stereo pair. Compared to
    traditional techniques, learned models capture more fine-grained features and
    generalise better across varying lighting and texture conditions, though at
    the cost of higher computational requirements and dependence on large-scale
    training data.

    Two prominent monocular depth estimation models are MiDaS \citep{birkl2023midas}
    and Depth Anything \citep{yang2024depth}. MiDaS employs transformer-based backbones
    (BEiT, Swin, SwinV2) trained on large-scale mixed datasets with a scale-and-shift-invariant
    loss, yielding strong zero-shot generalisation across datasets. Depth Anything
    further improves zero-shot performance by leveraging large volumes of unlabelled
    data and preserving semantic priors from pre-trained encoders; it
    consistently outperforms MiDaS on unseen benchmarks, particularly under limited
    computational budgets.
    \begin{figure}[htbp]
        \centering
        \subfigure[MiDaS, 1\,m]{\includegraphics[width=0.3\textwidth]{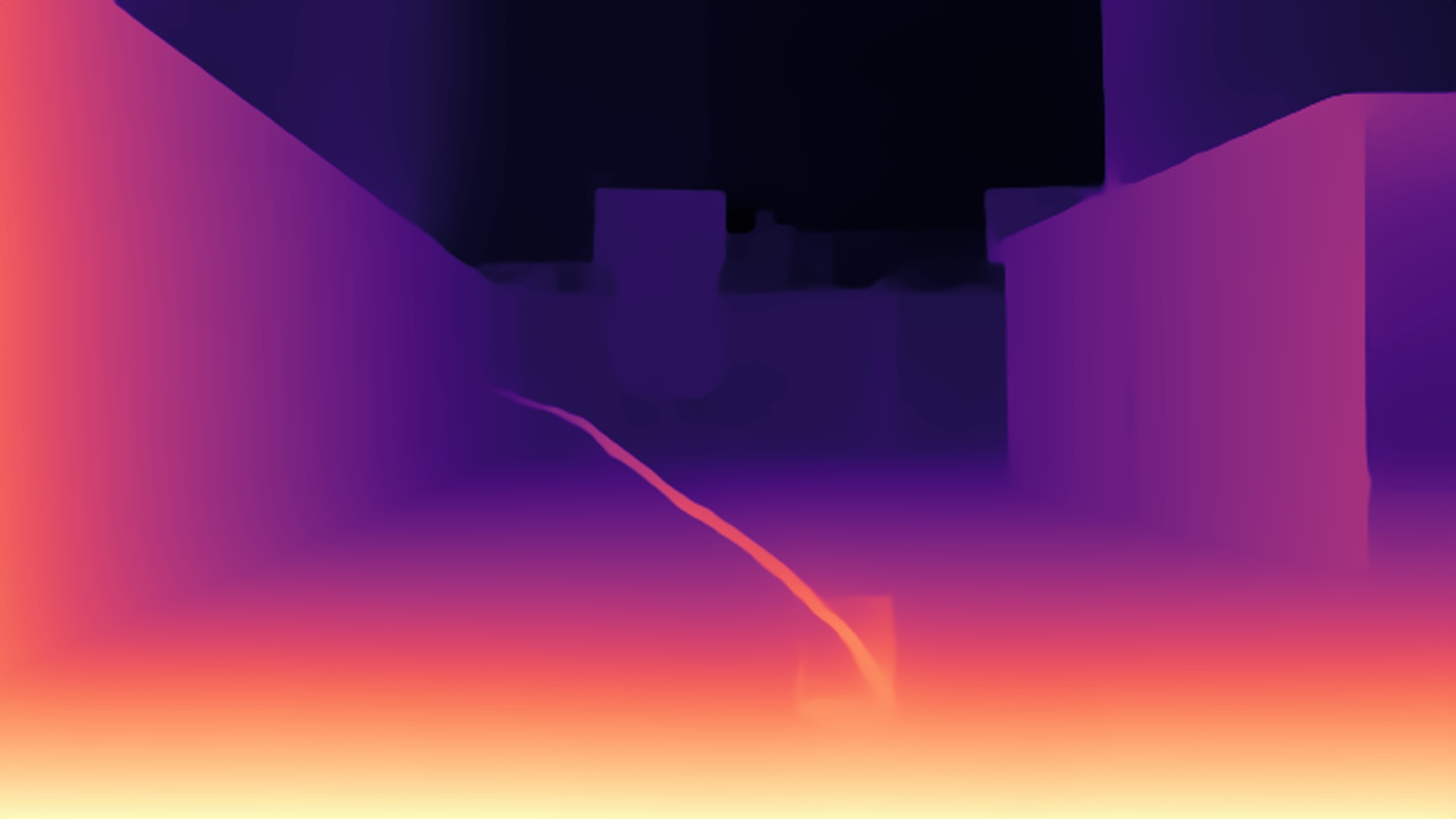}}
        \subfigure[Depth Anything, 1\,m]{\includegraphics[width=0.3\textwidth]{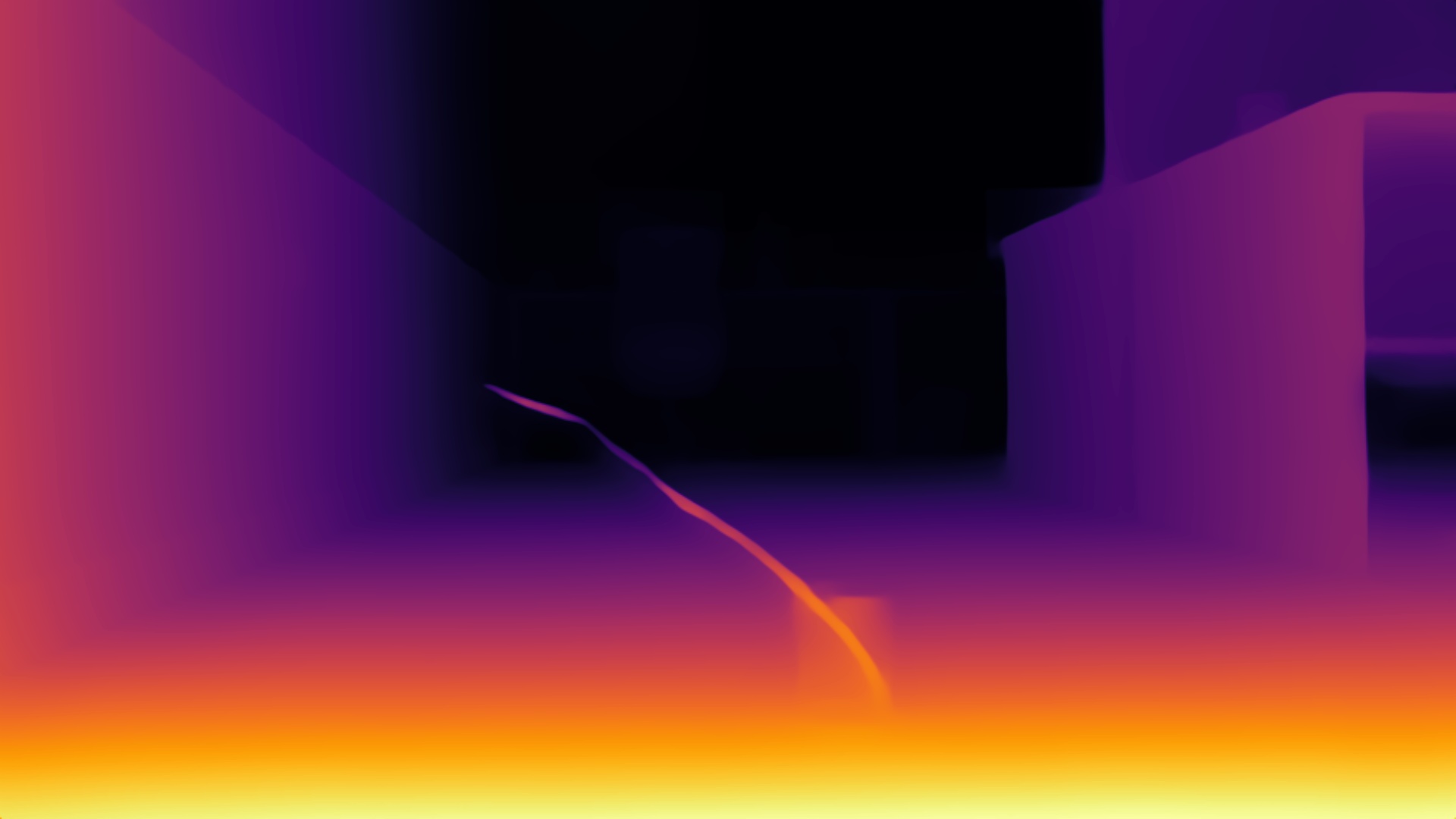}}
        \subfigure[MiDaS, 1.5\,m]{\includegraphics[width=0.3\textwidth]{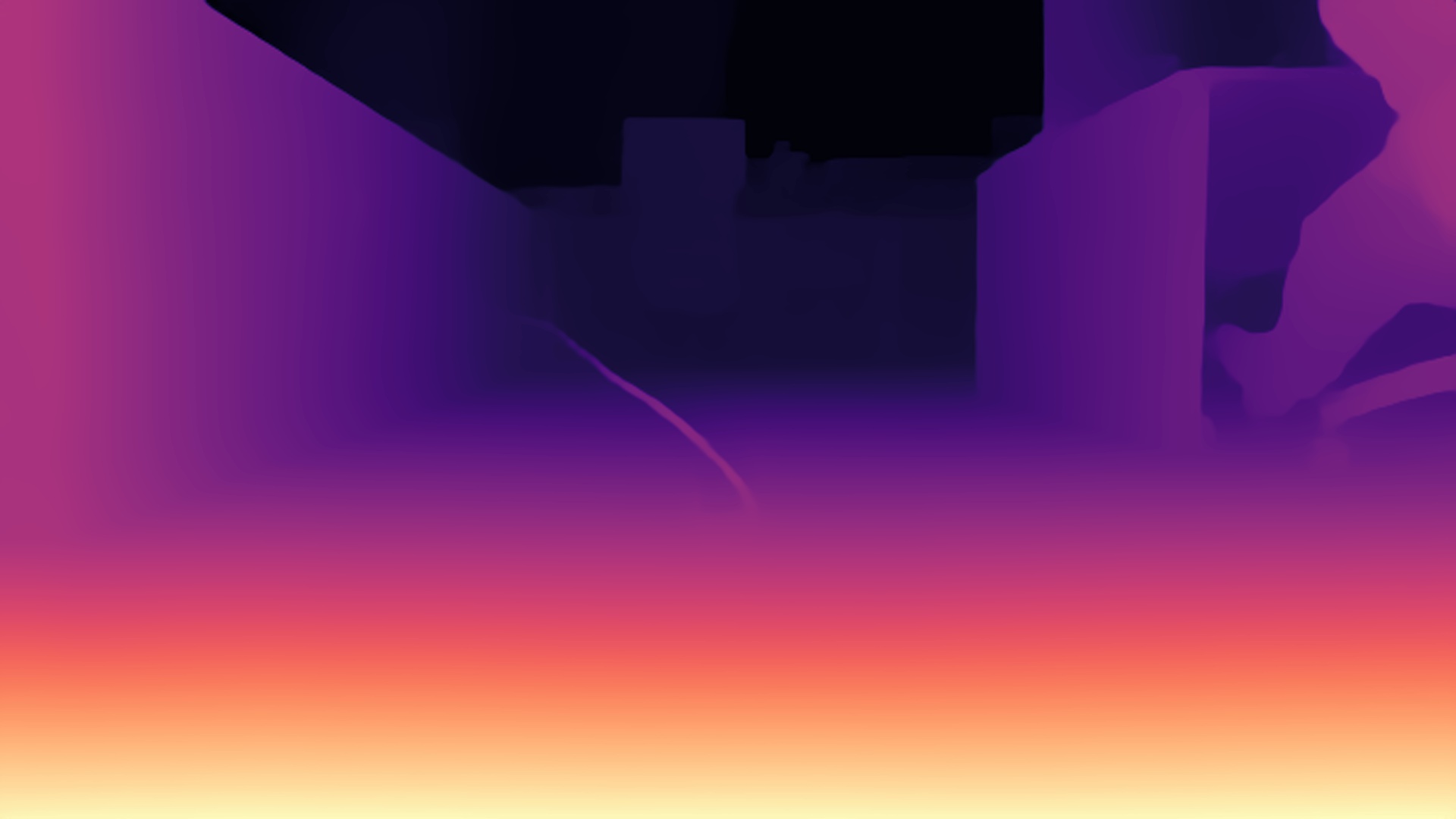}}
        \subfigure[Depth Anything, 1.5\,m]{\includegraphics[width=0.3\textwidth]{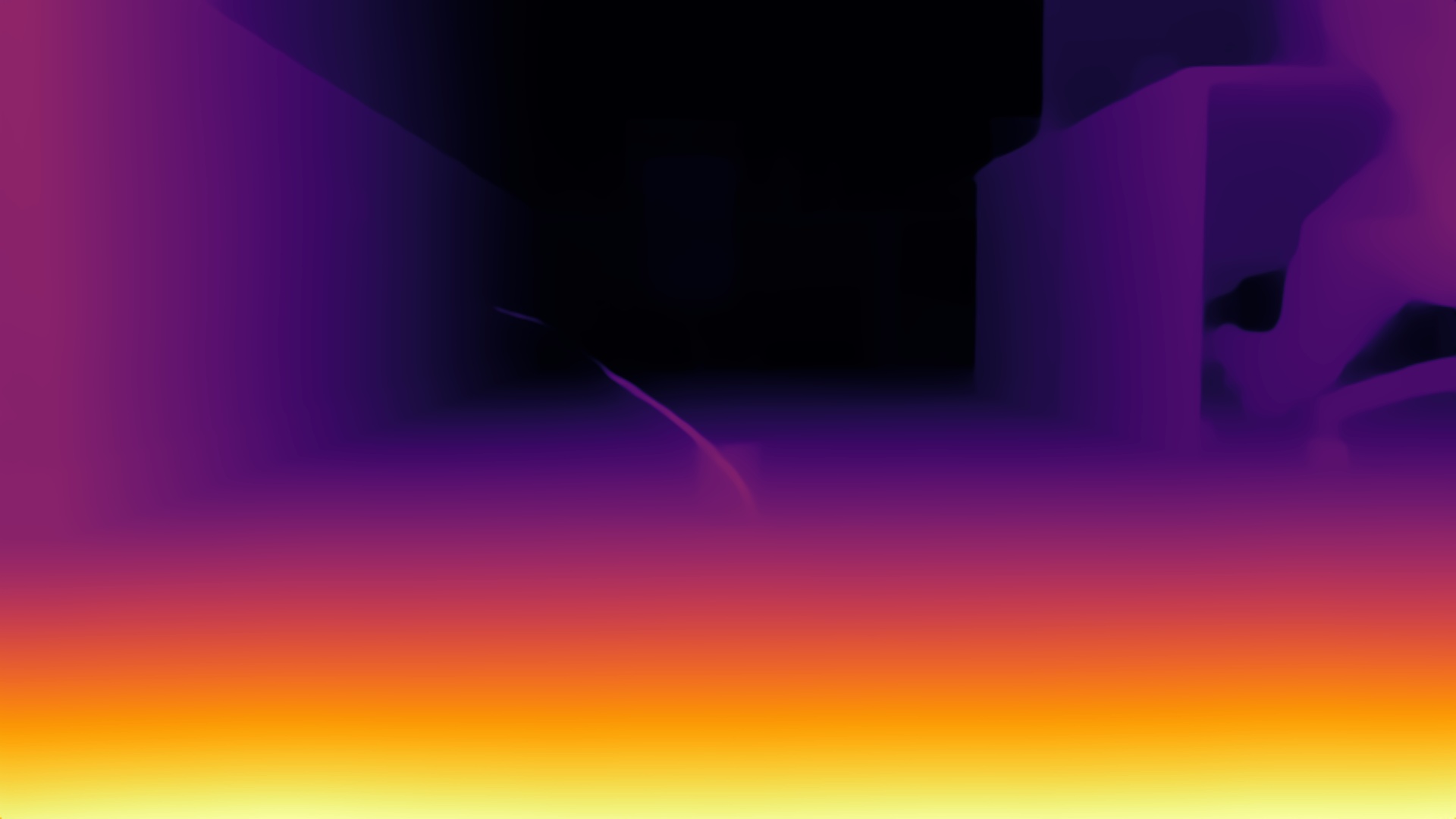}}
        \vspace{\baselineskip}
        \subfigure[MiDaS, 2\,m]{\includegraphics[width=0.3\textwidth]{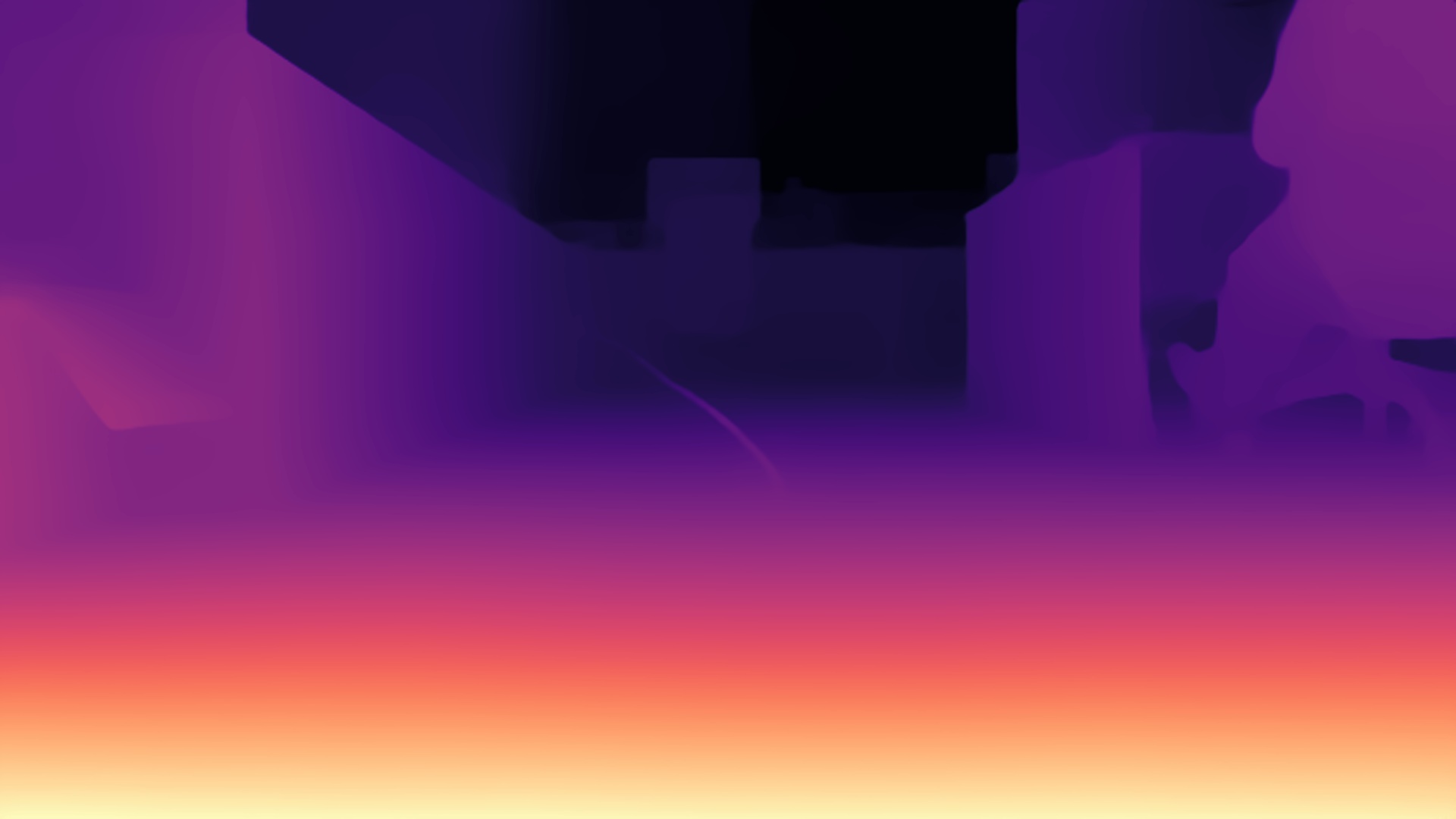}}
        \subfigure[Depth Anything, 2\,m]{\includegraphics[width=0.3\textwidth]{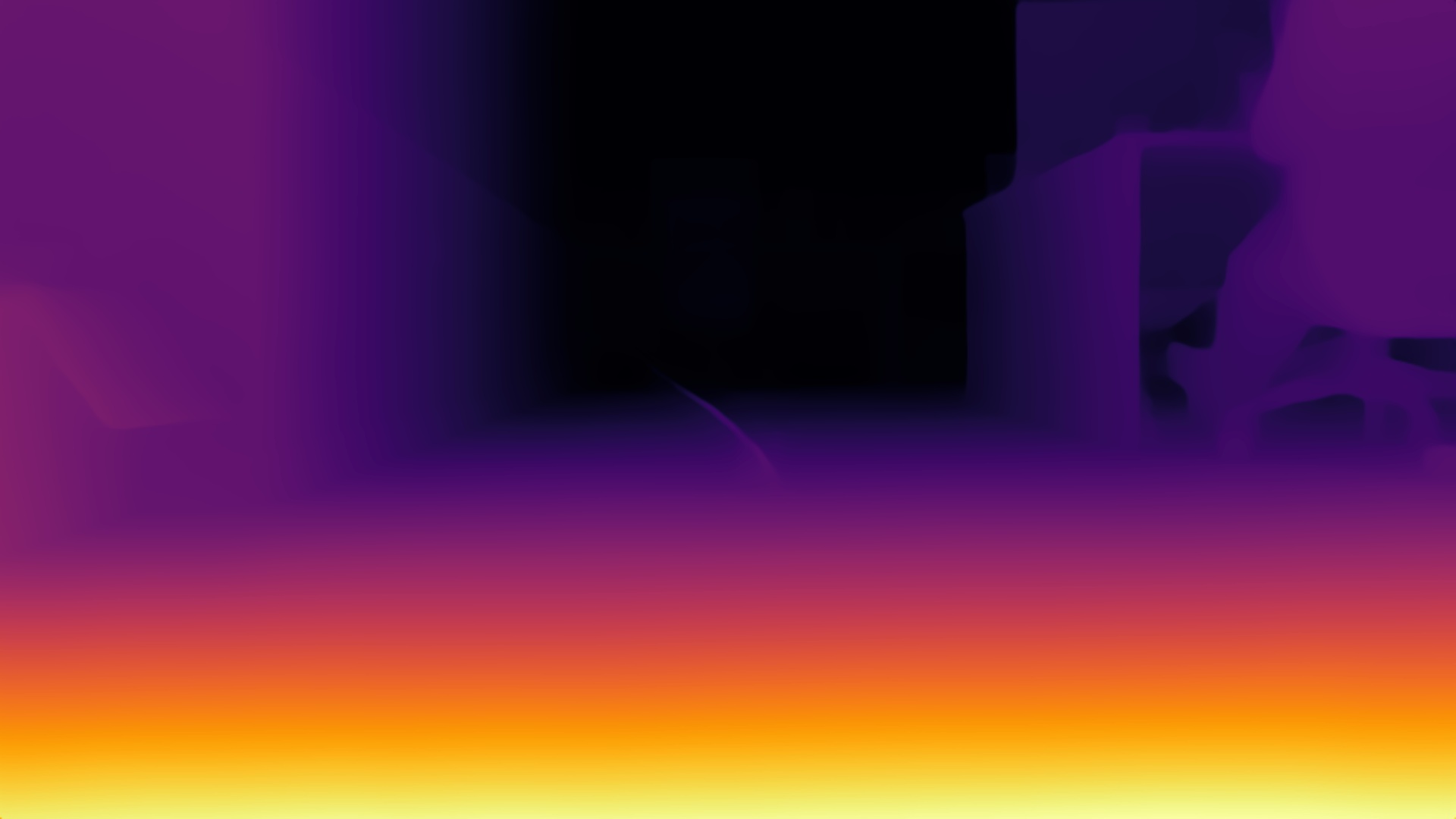}}
        \caption{Comparison of depth maps generated by MiDaS and Depth Anything
        models at branch distances of 1m, 1.5m, and 2m from the camera}
        \label{fig:monocular_comparison}
    \end{figure}

    We chose to use stereo depth estimation because it is generally more accurate
    than monocular depth estimation. Binocular depth estimation calculates the
    parallax between pixels by comparing two images taken from different angles
    to derive depth information, which provides a more accurate and detailed depth
    map.

    Commonly used methods for binocular depth prediction include ACVNet \citep{xu2022attention},
    GWCNet \citep{guo2019group}, PSMNet \citep{chang2018pyramid}, and RAFT-Stereo
    \citep{lipson2021raft}. We adopt NeRF-Supervised Deep Stereo \citep{tosi2023nerf}
    for depth estimation due to its capability to generate high-quality depth
    maps without the necessity for ground truth depth annotations. This approach
    synthesizes stereoscopic training data by utilizing Neural Radiance Fields.
    The methodology proceeds as follows:

    A handheld camera captures a sequence of images from a static scene, and COLMAP
    estimates the intrinsic parameters and camera pose for each view. A NeRF
    model trained on this data can then synthesise novel views from arbitrary
    camera perspectives, from which virtual stereo image pairs are rendered to train
    the depth estimation network. A third rendered image helps mitigate
    occlusion artifacts, further improving estimation accuracy.

    Beyond monocular approaches, deep learning has also been applied to multi-view
    disparity map prediction. In this research, we take binocular stereo vision
    as an example and conduct detailed tests and comparisons of different stereo
    matching architectures. Specifically, five different architectures, PSMNet,
    ACVNet, GWCNet, MobileStereoNet, and RAFT-Stereo, were selected and fine-tuning
    tests were conducted on different datasets.

    In the experiments, we have selected several common stereo matching datasets,
    including KITTI and Scene Flow datasets. Each architecture is fully fine-tuned
    on these datasets to ensure its adaptability and stability in different scenarios.

    In this study, we investigate the effect of fine-tuning on disparity map
    prediction across different datasets using PSMNet as an example. We selected
    three different datasets: KITTI 2012, KITTI 2015, and Scene Flow, and conducted
    detailed fine-tuning and testing on PSMNet. We trained on the KITTI 2012 and
    KITTI 2015 datasets for 100 epochs each. These two datasets mainly contain stereo
    image pairs of real-world city street scenes with high quality ground truth data.
    During the fine-tuning process, we found that although there are some differences
    in the scenes and labelling between these two datasets, training on the same
    dataset for a long period of time can significantly improve the prediction
    accuracy of the model. After PSMNet is fully trained on these two datasets,
    its prediction quality improves noticeably across all metrics; the disparity
    estimation error on the KITTI dataset drops considerably with 100-epoch training,
    indicating stronger dataset-specific learning. Applying these KITTI-trained models
    to our own tree-branch dataset, however, yields less satisfactory results.
    This phenomenon suggests that there are significant differences in data
    distribution and features between the two datasets; the KITTI dataset mainly
    contains urban streetscapes, whereas our own dataset mainly contains tree
    branches in natural scenes, which are very different in terms of scene and object
    features. Therefore, the model trained only on the KITTI dataset cannot be generalised
    and adapted to different types of scenes (e.g. tree branches).

    \begin{figure}[htbp]
        \centering
        \subfigure[original left image]{\includegraphics[width=0.3\textwidth]{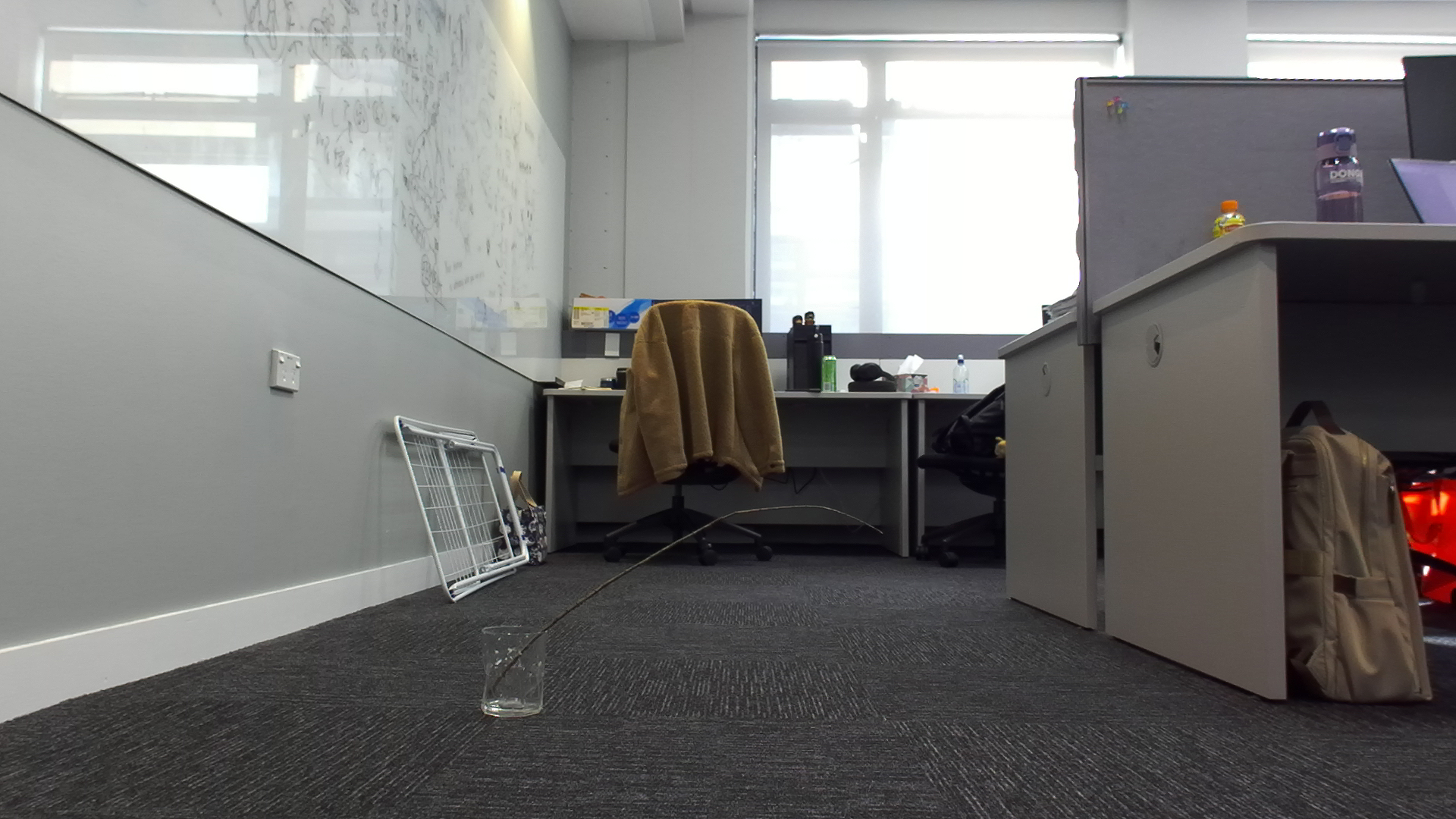}}
        \subfigure[Scene Flow pre-trained]{\includegraphics[width=0.3\textwidth]{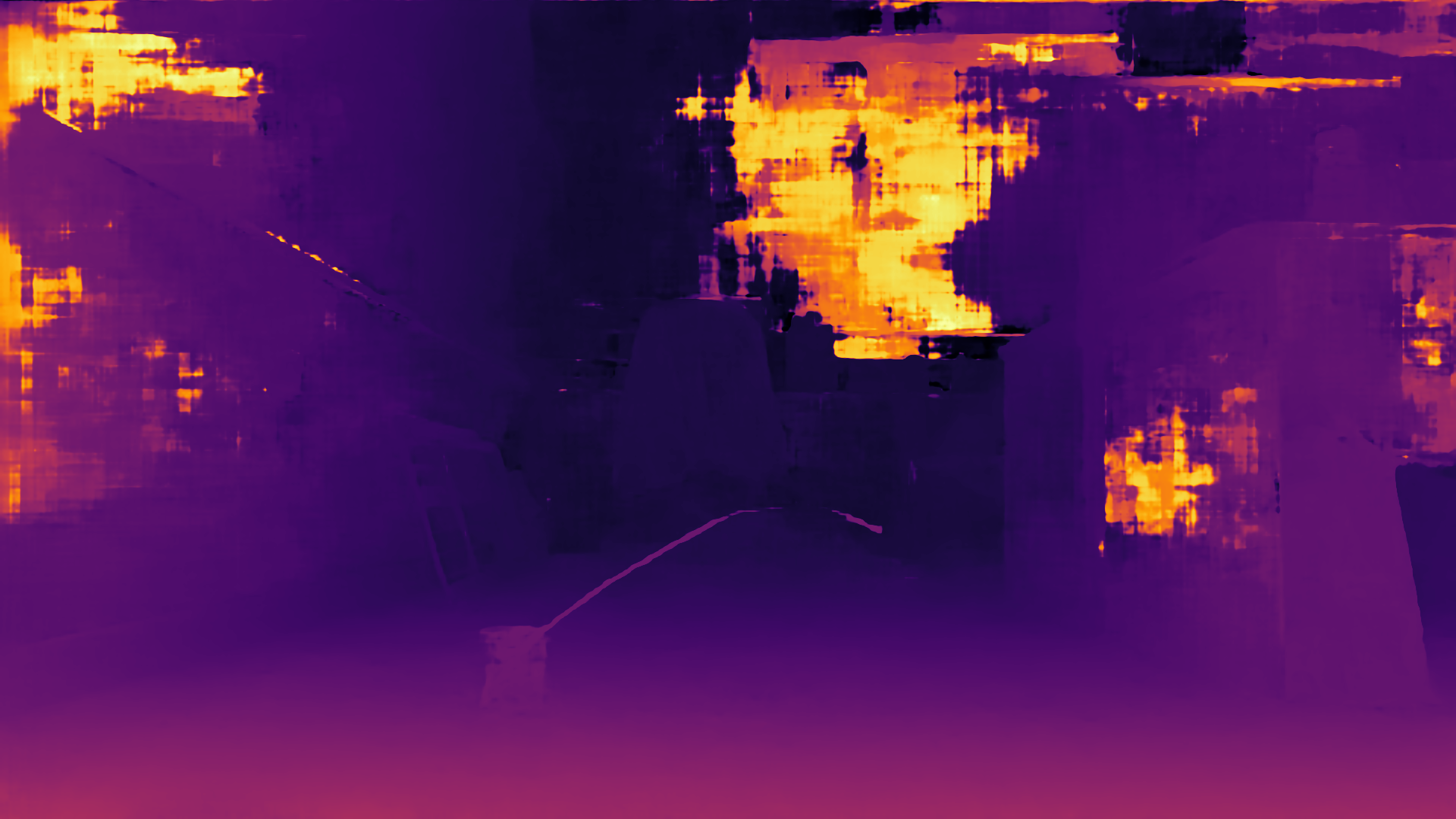}}
        \subfigure[KITTI 2012 pre-trained]{\includegraphics[width=0.3\textwidth]{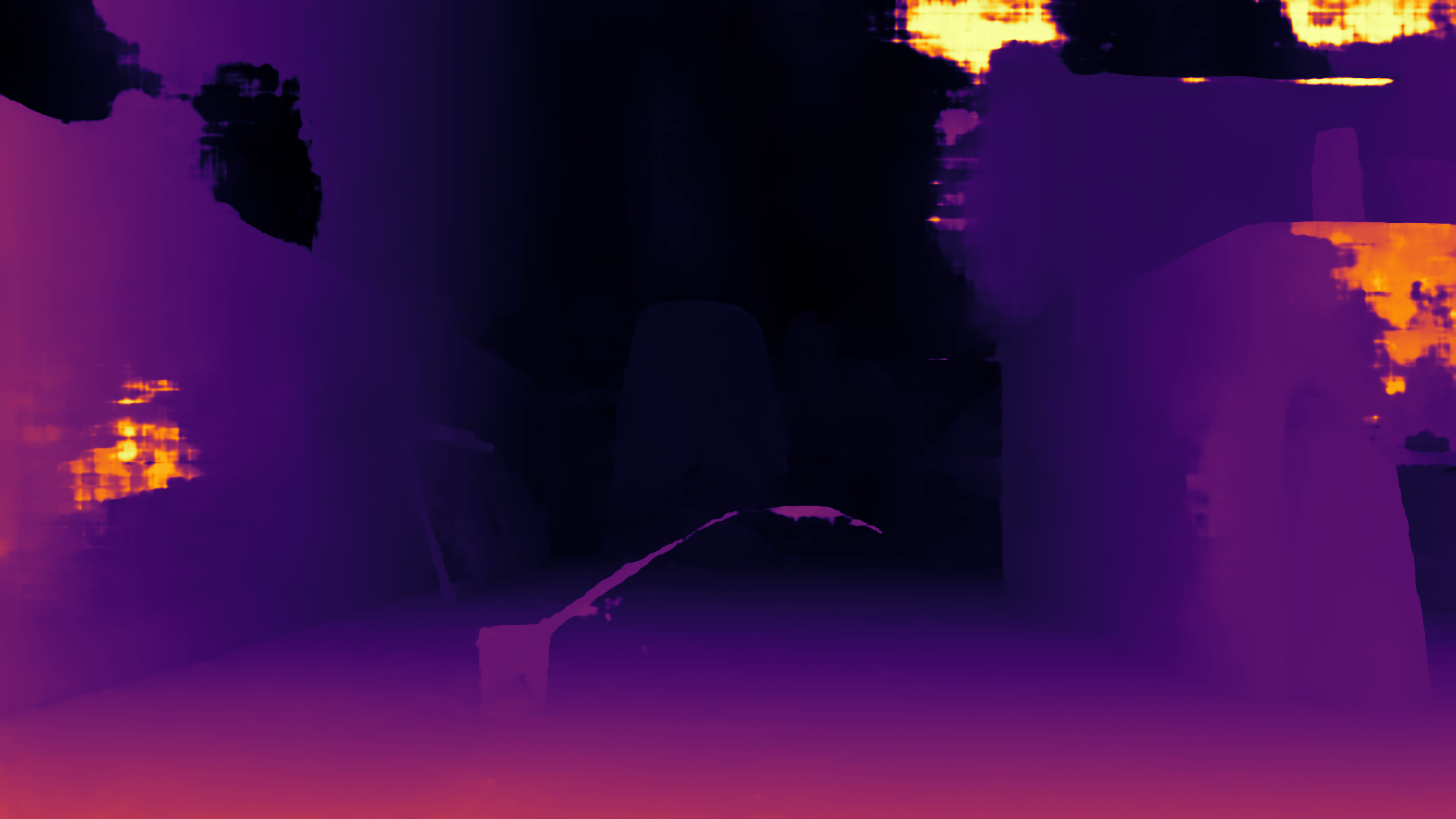}}
        \subfigure[KITTI 2012, 100 epochs]{\includegraphics[width=0.3\textwidth]{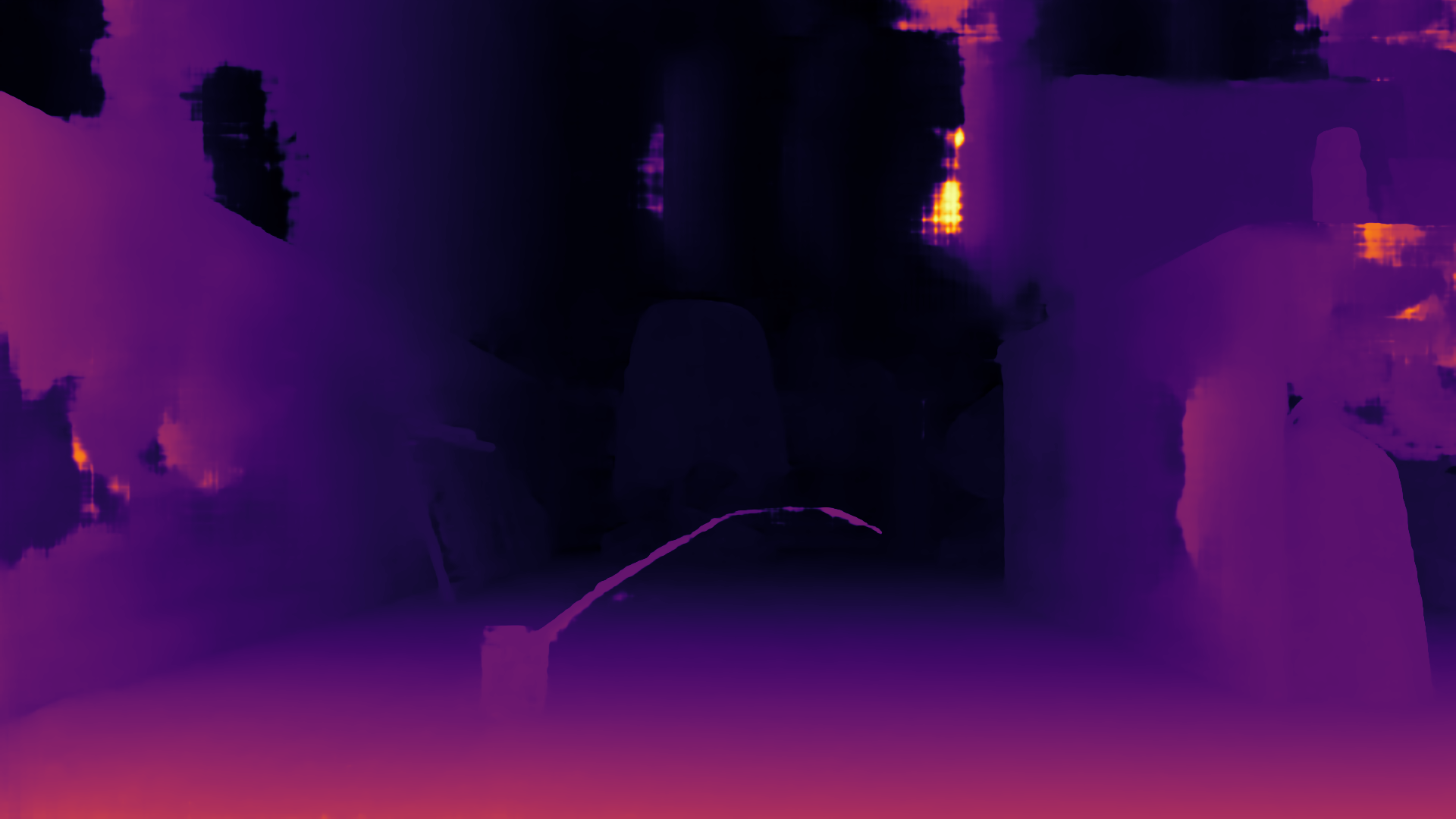}}
        \subfigure[KITTI 2015 pre-trained]{\includegraphics[width=0.3\textwidth]{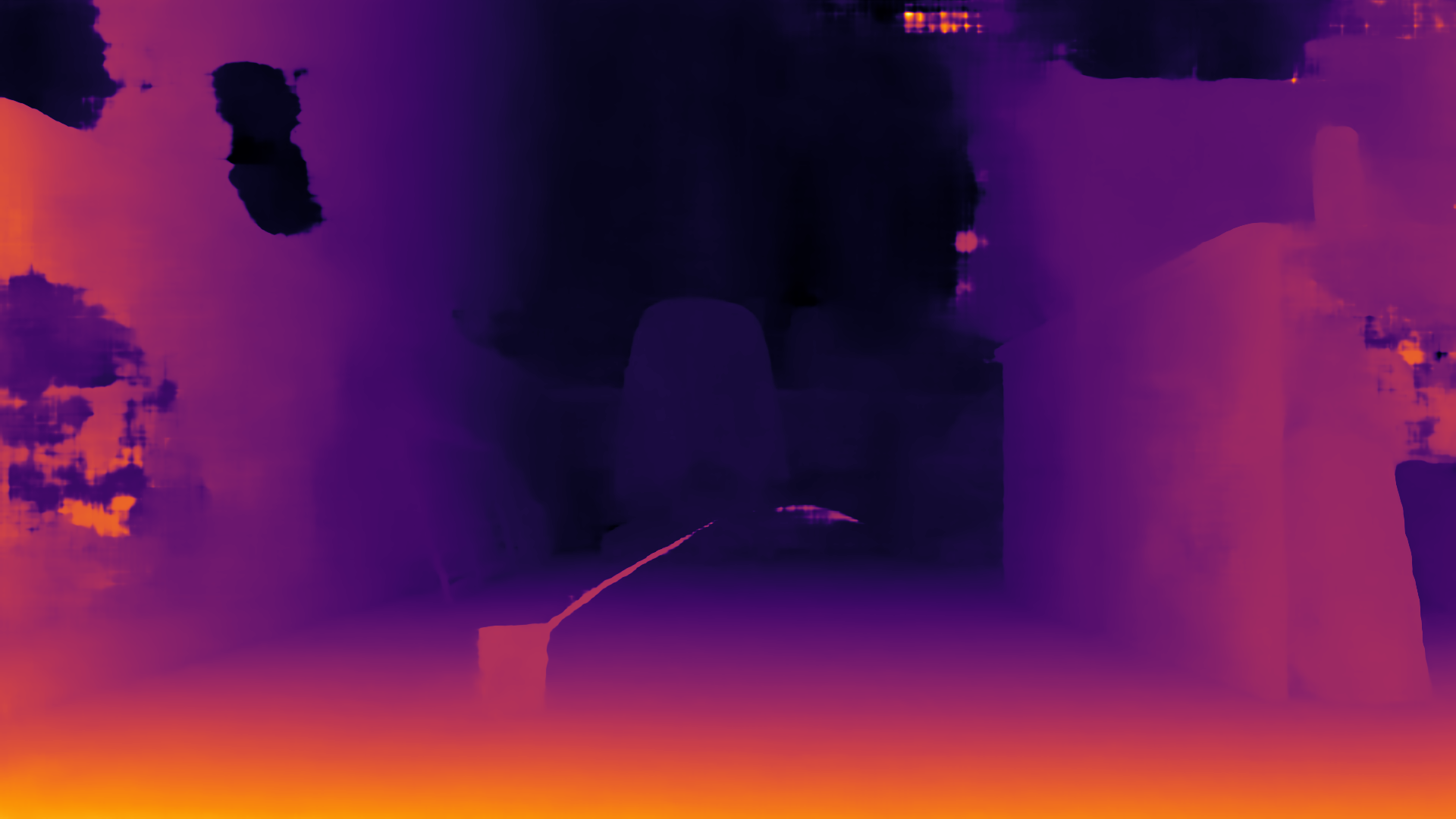}}
        \subfigure[KITTI 2015, 100 epochs]{\includegraphics[width=0.3\textwidth]{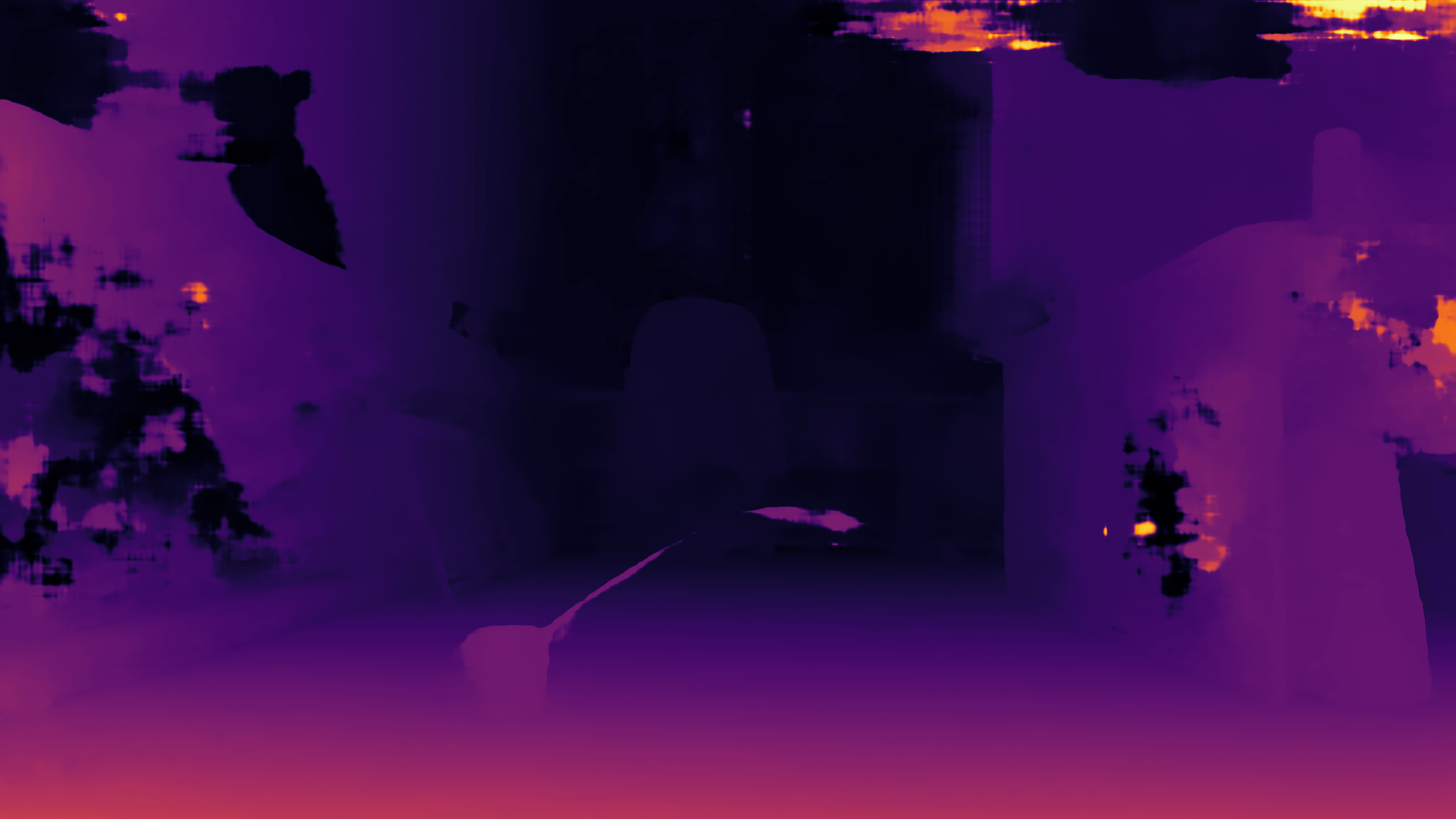}}
        \caption{Comparison of PSMNet Fine-tuning Results Across Different
        Pretrained Models and Training Epochs }
        \label{fig:psmnet_finetuning}
    \end{figure}

    The visual comparison of disparity maps produced by each architecture on the
    same stereo input pair is presented below.

    \begin{figure}[htbp]
        \centering
        \subfigure[PSMNet]{\includegraphics[width=0.3\textwidth]{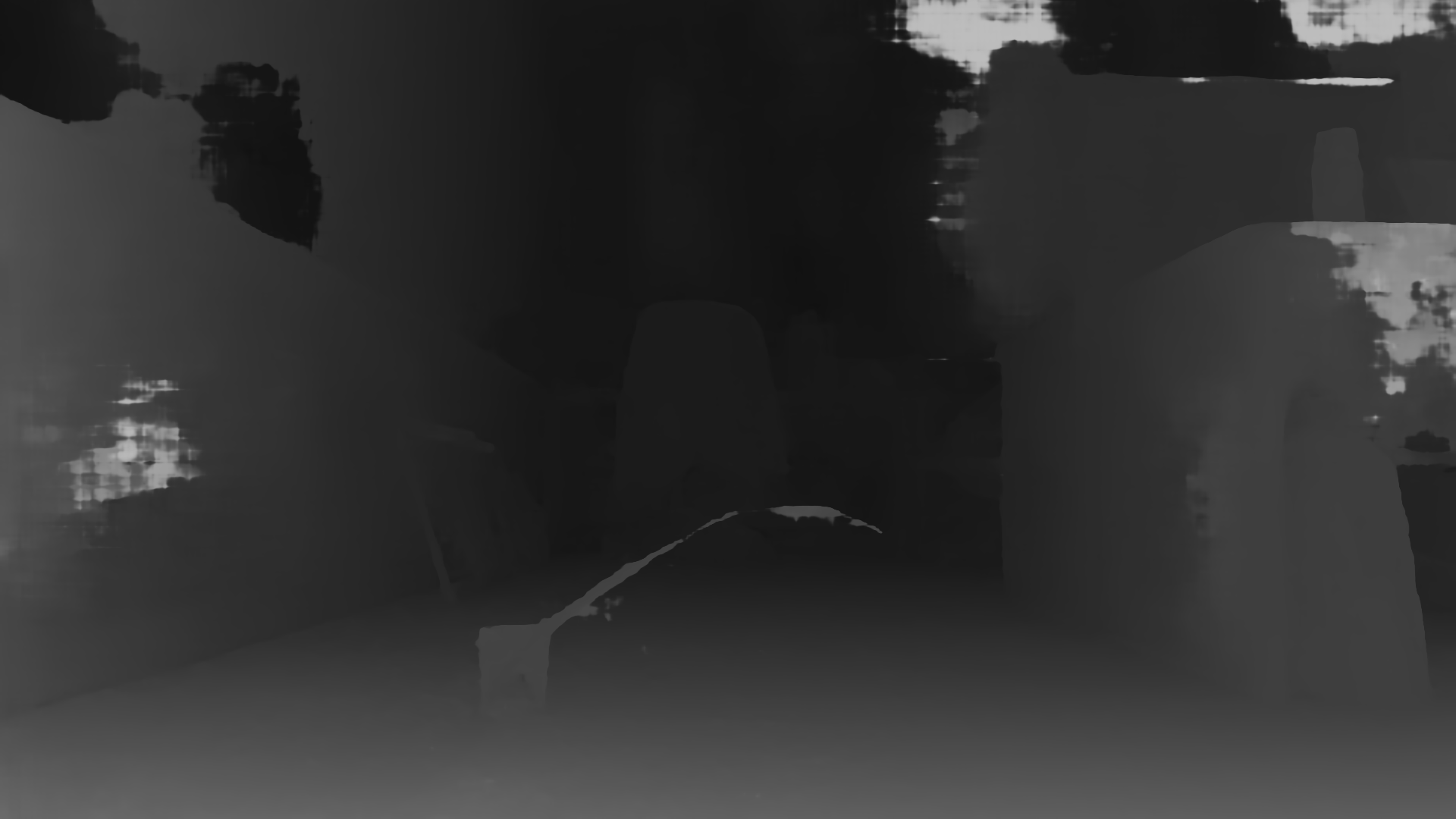}} \subfigure[ACVNet]{\includegraphics[width=0.3\textwidth]{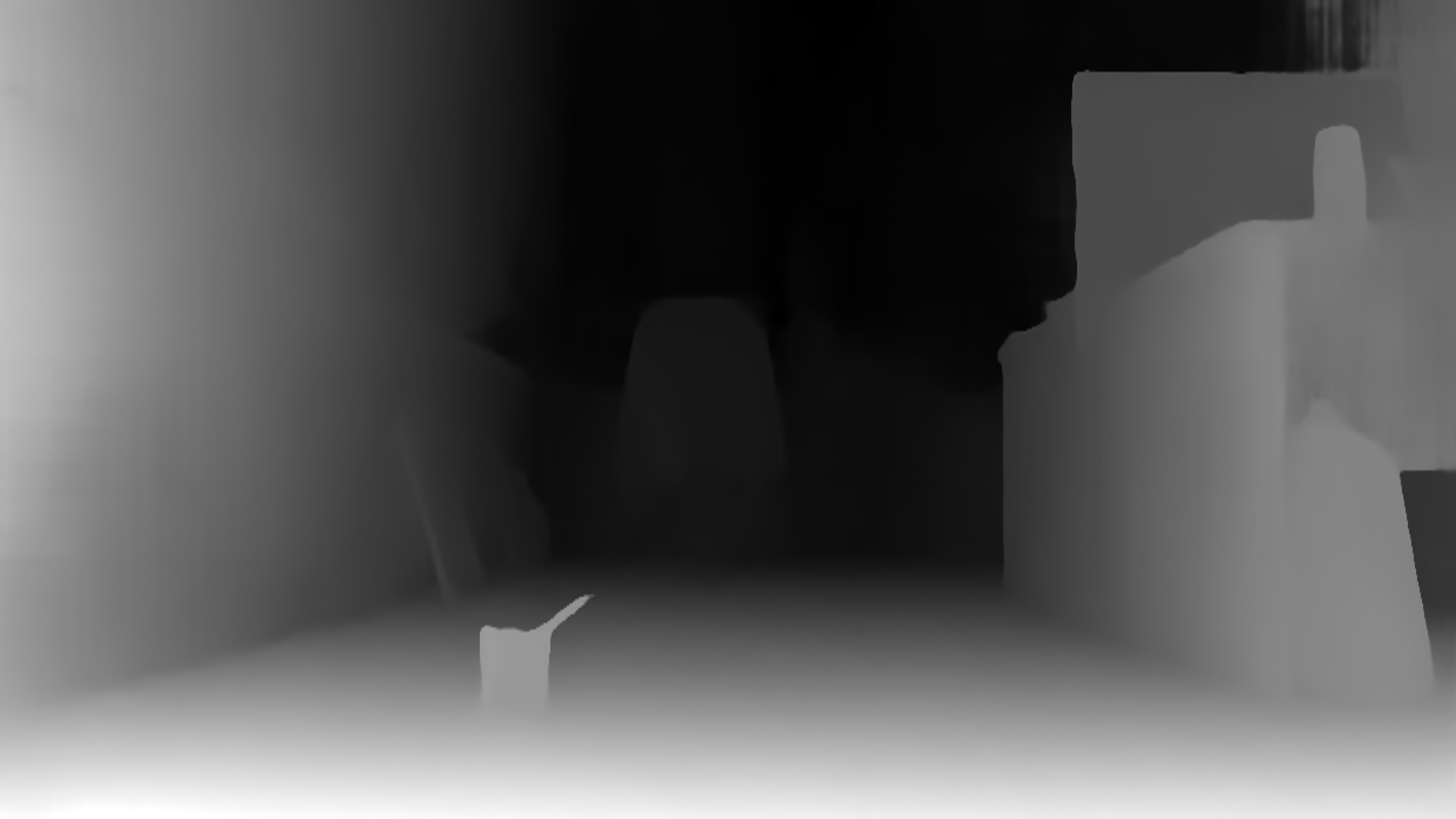}}
        \subfigure[GWCNet]{\includegraphics[width=0.3\textwidth]{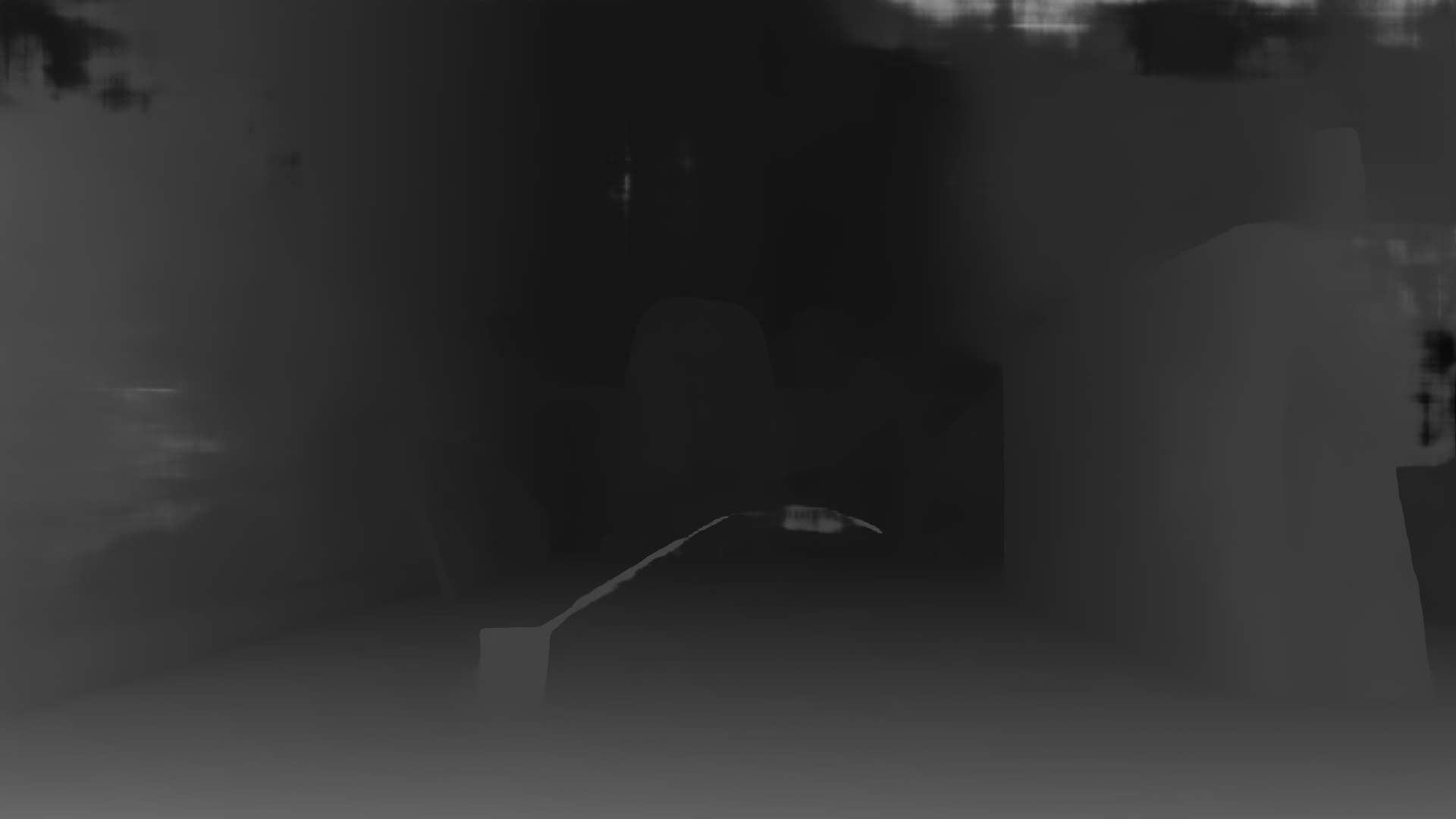}} \subfigure[MobileStereoNet]{\includegraphics[width=0.3\textwidth]{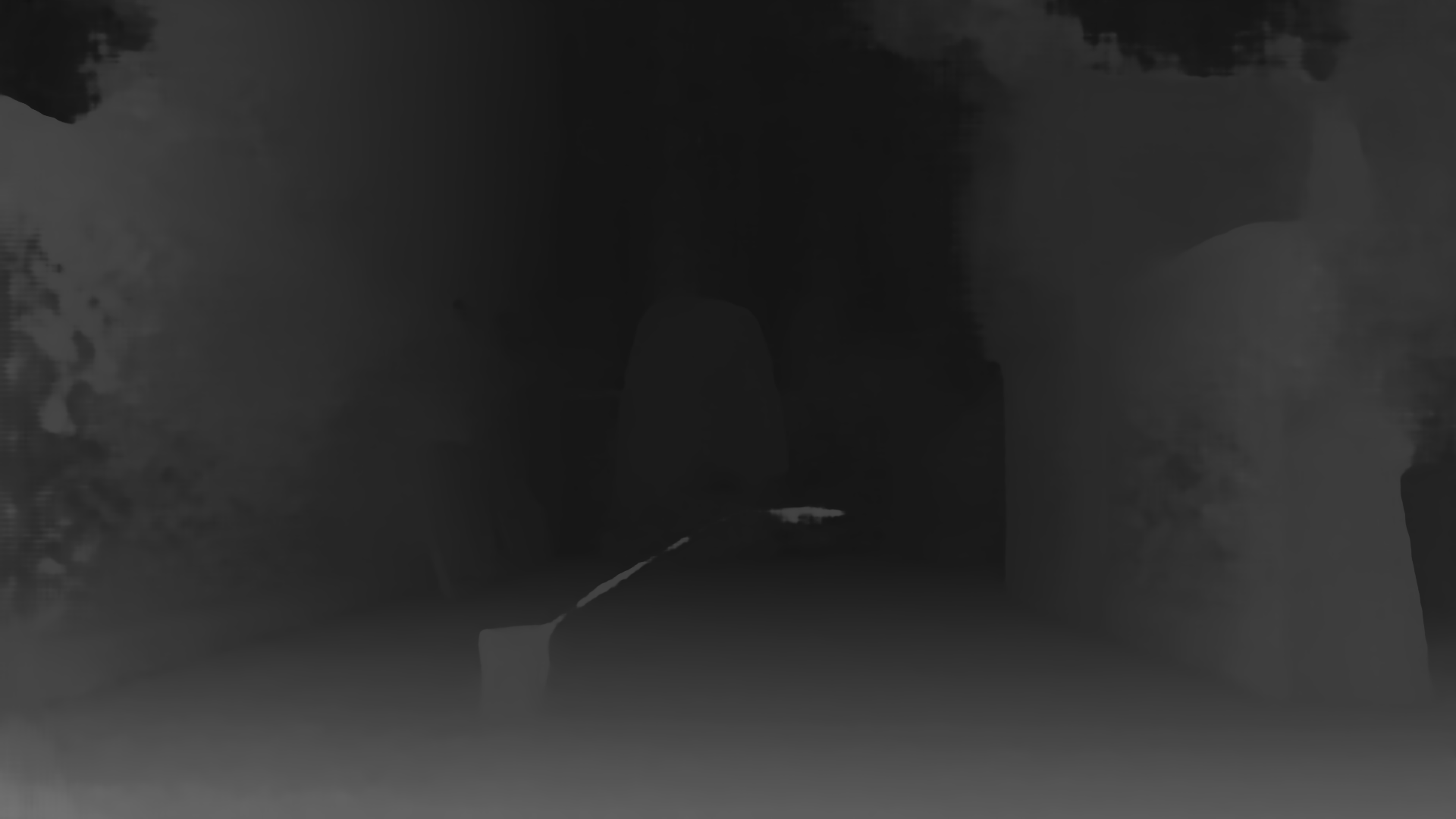}}
        \subfigure[RAFT-Stereo]{\includegraphics[width=0.3\textwidth]{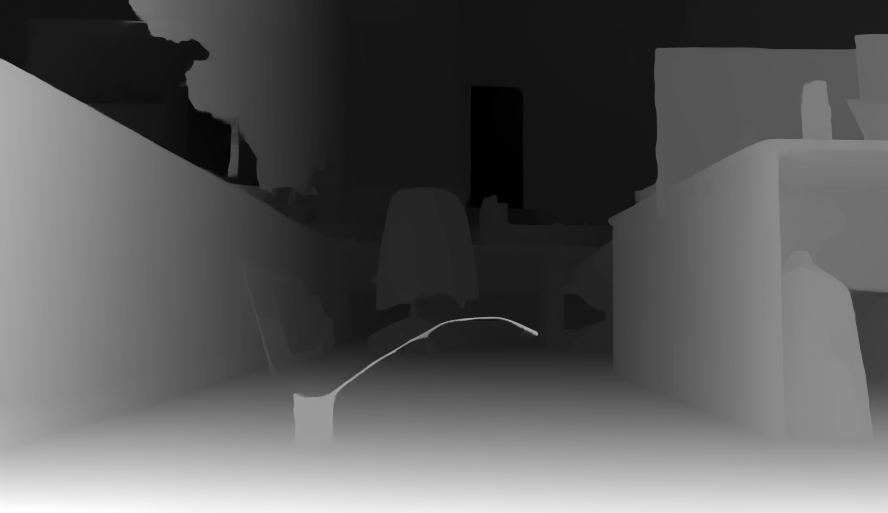}}
        \subfigure[NeRF-Stereo]{\includegraphics[width=0.3\textwidth]{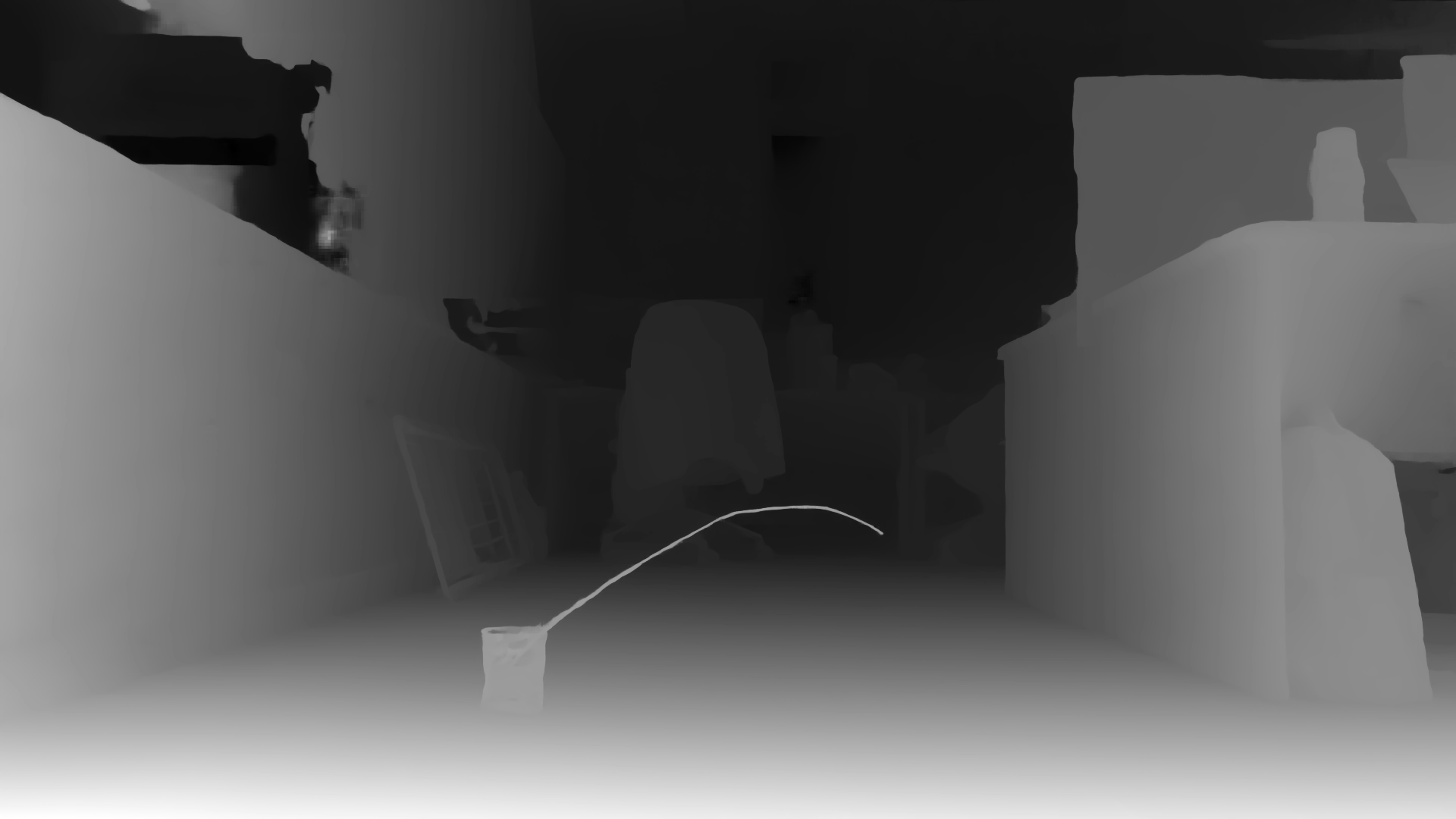}}
        \caption{Comparison of Stereo Matching Results Across Different Neural
        Network Models}
        \label{fig:stereo_comparison}
    \end{figure}
    \newpage
    \subsection{Combination}

    The remaining step is straightforward: combining branch detection with depth
    computation. Our goal is to detect the coordinates of points surrounding the
    branch, apply a triangulation-based method to locate points lying on the branch,
    and then retrieve the corresponding depth values from the disparity map. Anomalous
    depth readings are filtered out via a median-based approach, and the
    remaining values are averaged to obtain the final distance from the branch
    to the camera plane.\\
    Let $P$ denote the set of predicted points, where the $i$-th point is $p_{i}$,
    giving $P=\{p_{1}, p_{2},..., p_{n}\}$ with $0<i\leq n$ and
    $p_{i}=( x_{i},y_{i})$. Here $x_{i}$ is the horizontal coordinate of point $p
    _{i}$ in the depth map and $y_{i}$ is the vertical coordinate of point $p_{i}$
    in the depth map.

    Using the triangular method, we compute the centroid of every three
    predicted points; these centroids are expected to fall on or near the branch.
    Let the centroid set be $P'$, $P'= \{p'_{1}, p'_{2},...,p'_{k}\}$,
    $0<k\leq \frac{n}{3}$, $p'_{g}= (x'_{g}, y'_{g})$, where $x'_{g}$ is the
    horizontal coordinate of centroid $p'_{g}$ in the depth map and $y'_{g}$ is
    the vertical coordinate of centroid $p'_{g}$ in the depth map.
    \begin{figure}[htbp]
        \centering
        \subfigure[predicted points]{\includegraphics[width=0.45\textwidth]{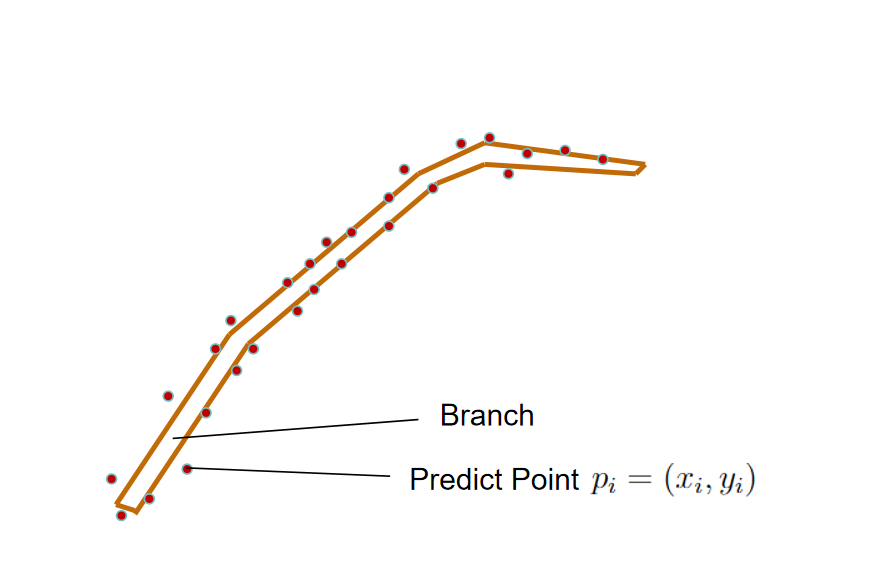}}
        \hfill \subfigure[grouping the closest points into triangles]{\includegraphics[width=0.35\textwidth]{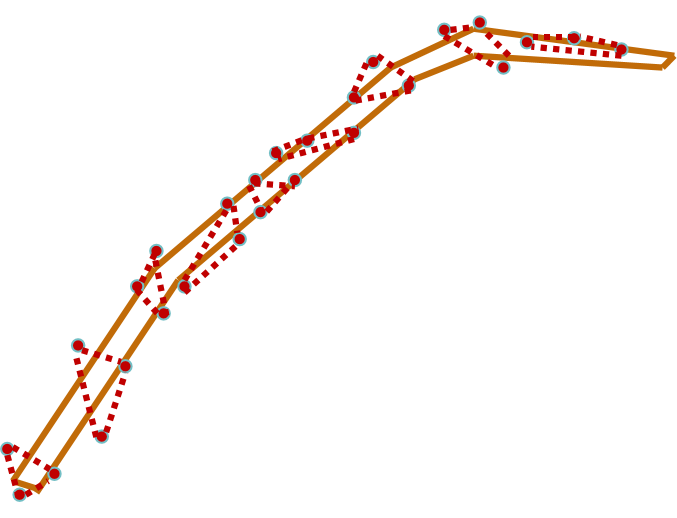}}
        \vspace{\baselineskip}
        \subfigure[computing centroids]{\includegraphics[width=0.45\textwidth]{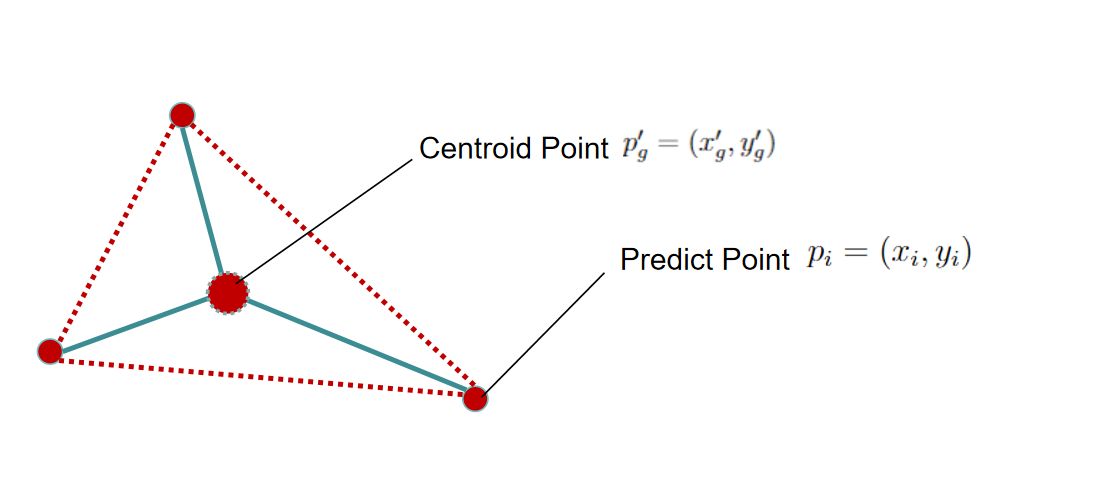}}
        \caption{Illustration of the centroid-based branch localisation process:
        (a) predicted points surrounding the branch, (b) grouping the closest
        points into triangles, and (c) computing their centroids.}
        \label{fig:centroid_process}
    \end{figure}

    Let each centroid be expanded by $m$ neighbouring sample points. The $j$-th point
    added around centroid $p'_{i}$ is denoted $p_{i,j}$. Let the expanded point
    set be $P''$:
    \[
        P'' = \{P'_{1}, P'_{2}, \ldots, P'_{n}\}
    \]where
    \[
        P''_{i}= \{p_{i,1}, p_{i,2}, \ldots, p_{i,m}\}
    \]and
    \[
        p_{i,j}= (x''_{i,j}, y''_{i,j}).
    \]Then we define the total points set as $P'''$:
    \[
        P''' = P'' + P'.
    \]MAD-based filtering is then applied: use MAD (Median Absolute Deviation) to
    remove depth values that deviate excessively from the median.

    \begin{equation}
        \text{MAD}= \text{median}(\lvert P''' - \text{median}(P''')\lvert)
    \end{equation}
    A point in $P'''$ is retained if its depth value falls within $\text{median}(
    P''')\pm k \cdot \text{MAD}$, where $k$ is a user-specified rejection
    threshold. The retained points form $\bar{P}$, and the final distance is
    computed as:
    \begin{equation}
        \text{distance}= \text{mean}(\bar{P}) \label{eq:distance}
    \end{equation}
    This yields the final estimated distance between the camera and the branches.
    The result is shown in the following figure, where all red dots represent the
    retained points in $\bar{P}$.

    This method is effective in reducing the amount of computation and at the same
    time obtaining the approximate distance between the branch and the camera more
    accurately. In addition, there is another method that can achieve the same result.
    Specifically, based on the predicted points, they are connected into
    irregular geometries and all points are extracted from them. These points are
    then processed as in Eq.~(\ref{eq:distance}), and the result is the approximate
    distance between the branch and the camera. The purpose of this is to obtain
    depth values for all points around the entire branch and itself. Due to the extremely
    large number of these points, it is possible to analyse them statistically. The
    combined results and analyses are presented in Section 4.

    \begin{figure}[htbp]
        \centering
        \subfigure[detected points on the depth map]{\includegraphics[width=0.45\textwidth]{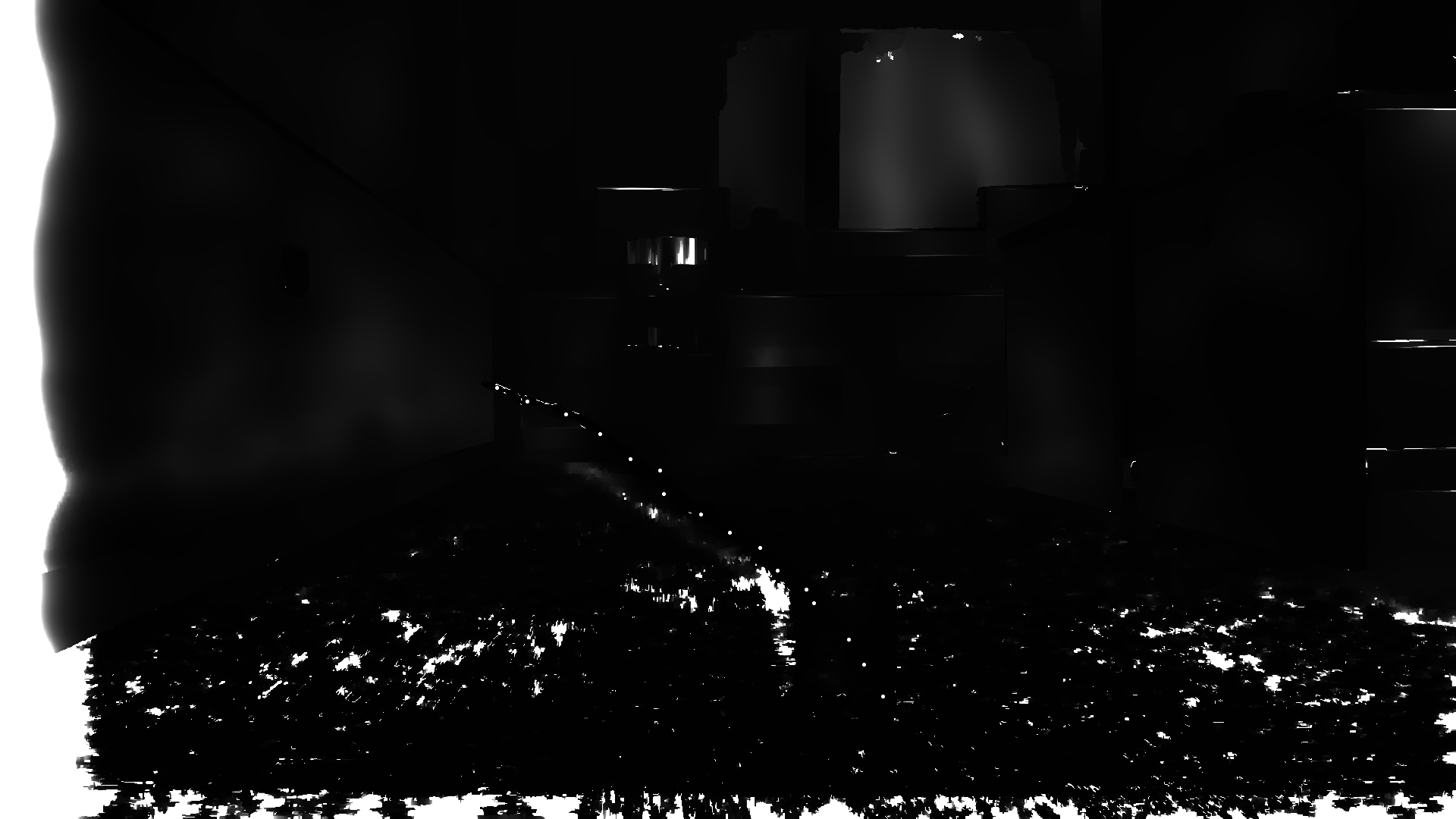}}
        \hfill \subfigure[detected points on the RGB image]{\includegraphics[width=0.45\textwidth]{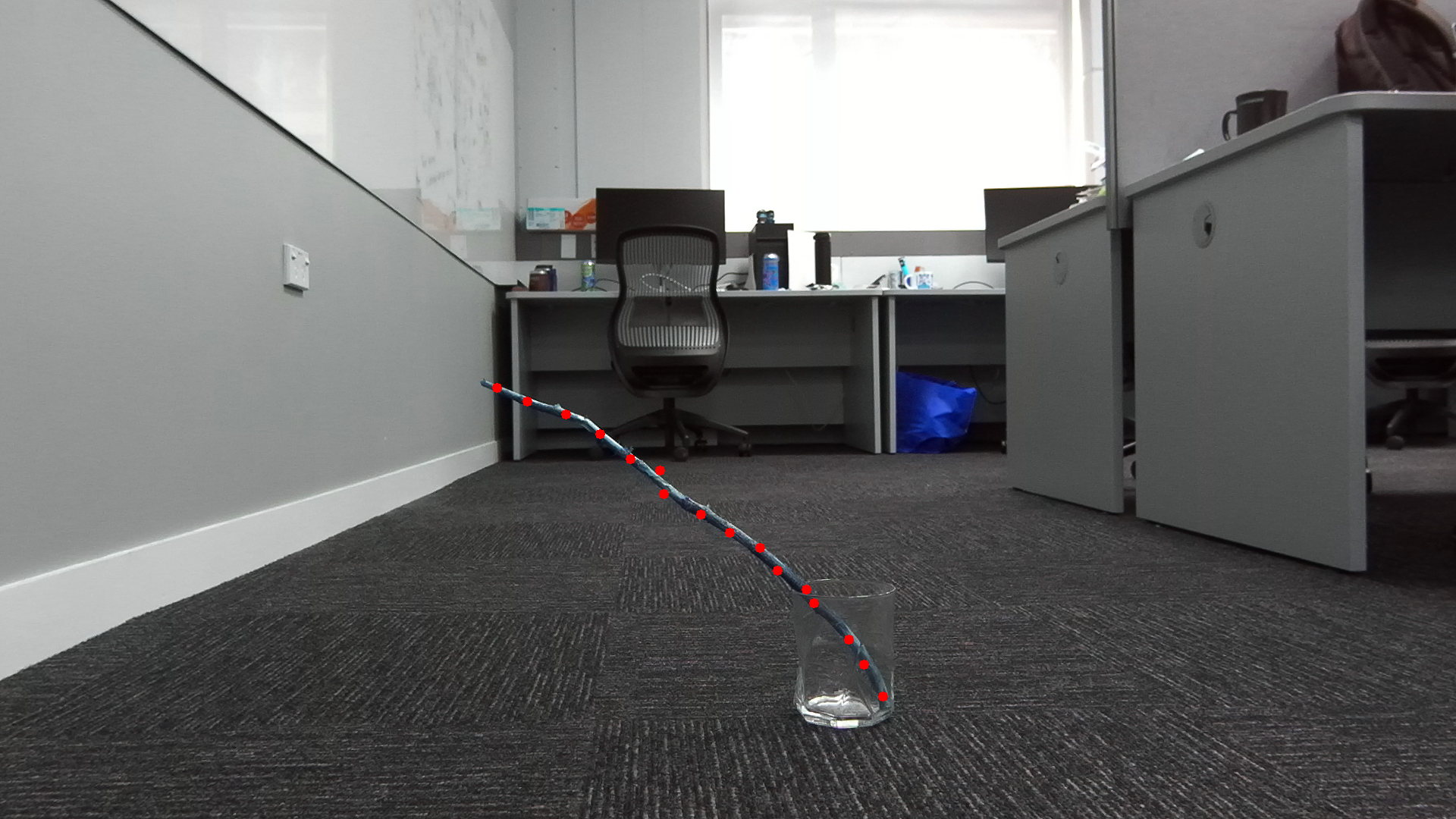}}
        \caption{Branch depth detection results: (a) detected points overlaid on
        the depth map using the centroid method, and (b) the corresponding
        points displayed on the RGB image.}
        \label{fig:centroid_depth}
    \end{figure}

    \section{Results and Analysis}
    This section demonstrates how the model obtained from YOLOv9e training can be
    used to generate depth maps using a combination of the traditional SGBM method
    as well as the NeRF technique based on the RAFT architecture to detect and determine
    the branch distances and compare them. Finally, the depth information of the
    branches is statistically analysed and the results are shown below.

    \begin{figure}[htbp]
        \centering
        \subfigure[SGBM depth 1m]{\includegraphics[width=0.3\textwidth]{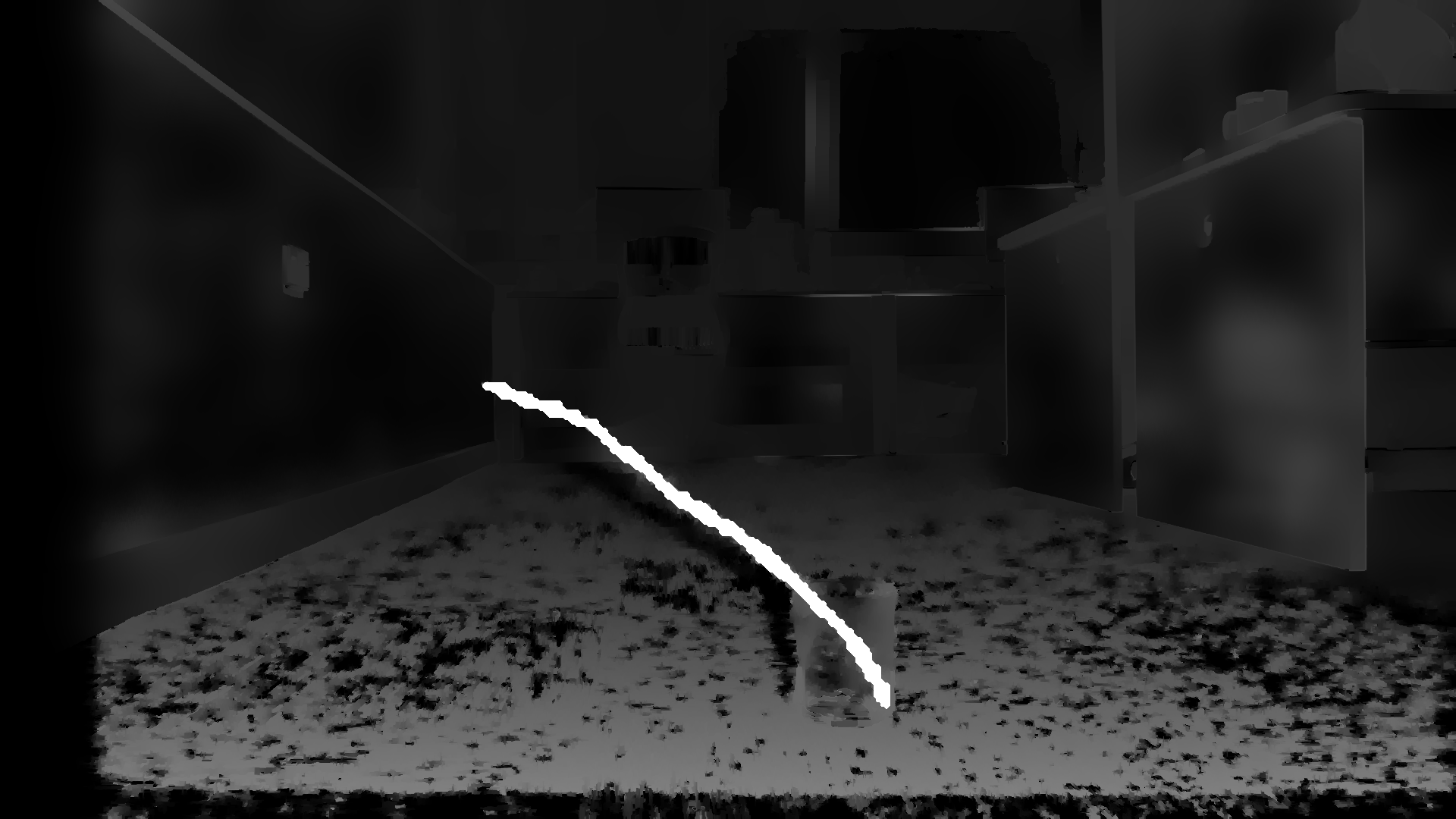}}
        \subfigure[SGBM depth 1.5m]{\includegraphics[width=0.3\textwidth]{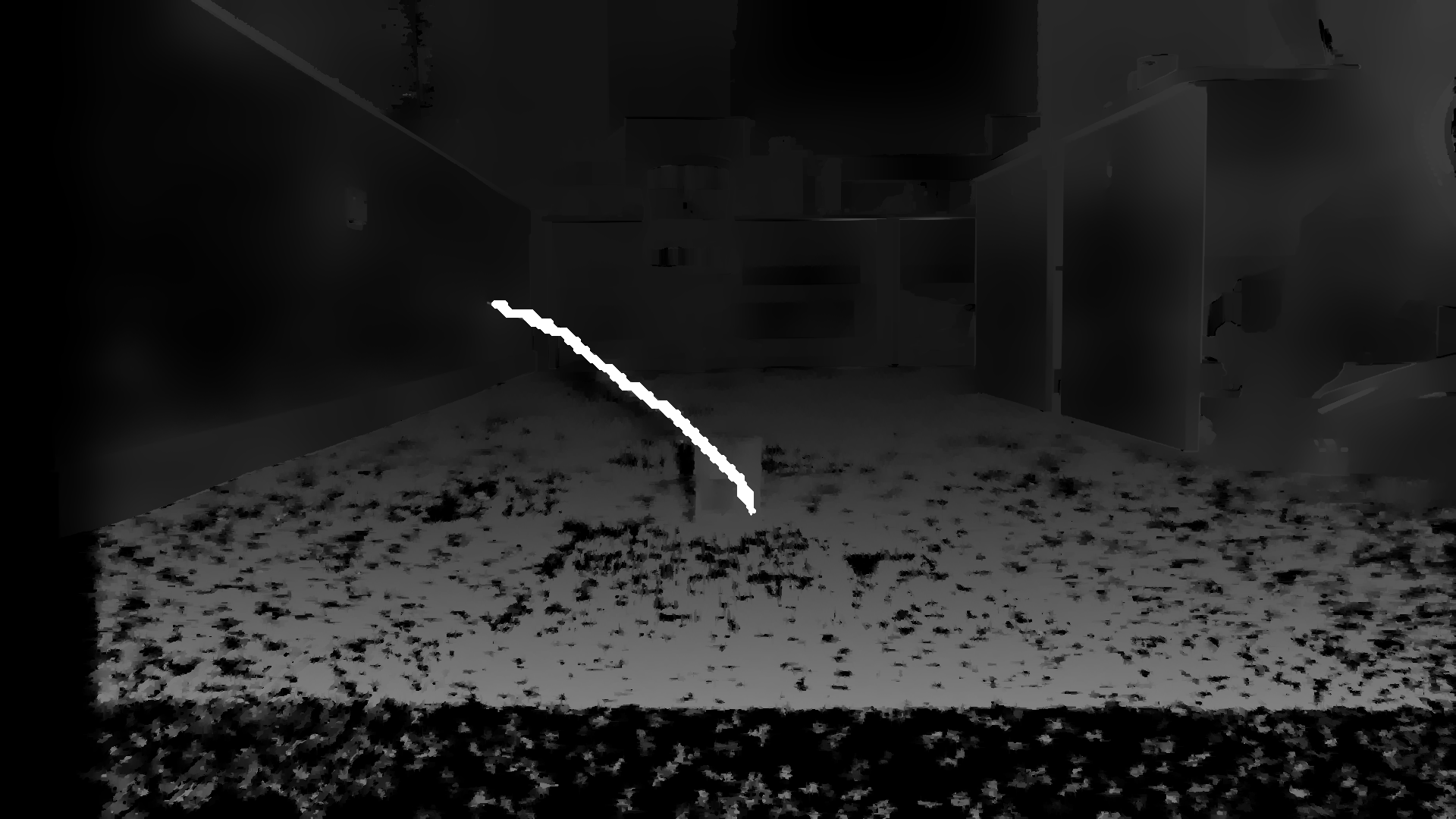}}
        \subfigure[SGBM depth 2m]{\includegraphics[width=0.3\textwidth]{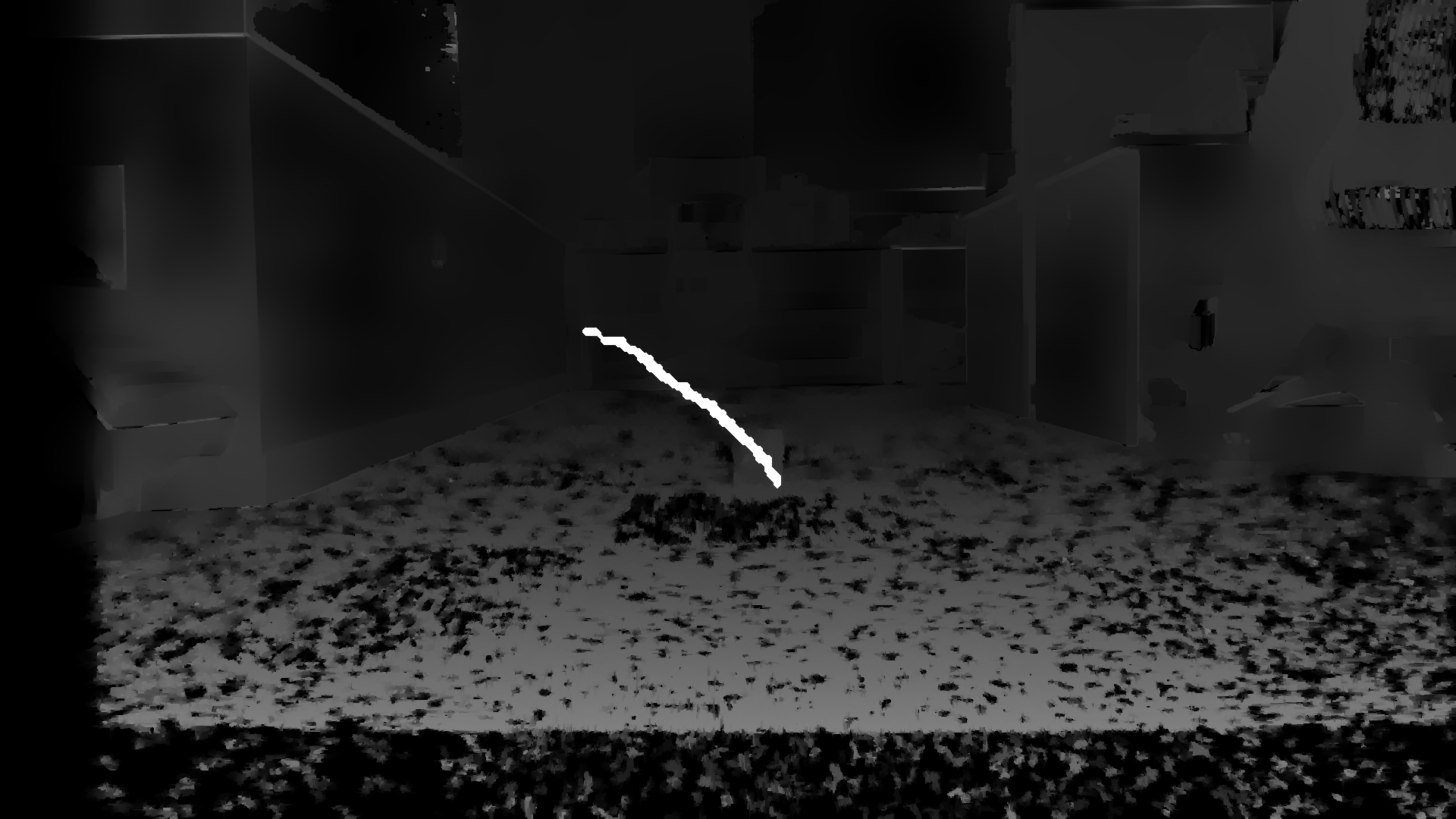}}
        \subfigure[NeRF-Stereo depth 1m]{\includegraphics[width=0.3\textwidth]{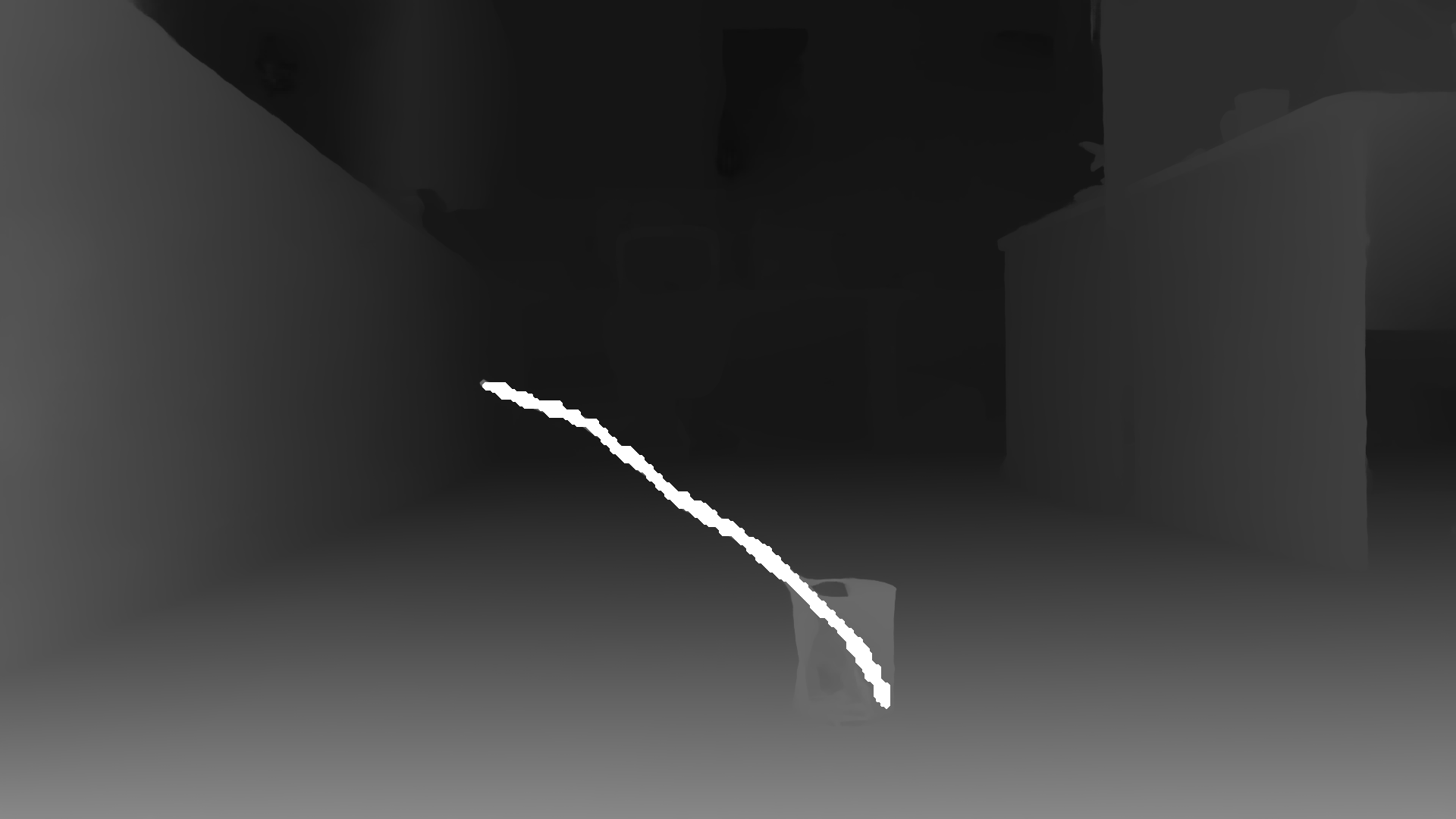}}
        \subfigure[NeRF-Stereo depth 1.5m]{\includegraphics[width=0.3\textwidth]{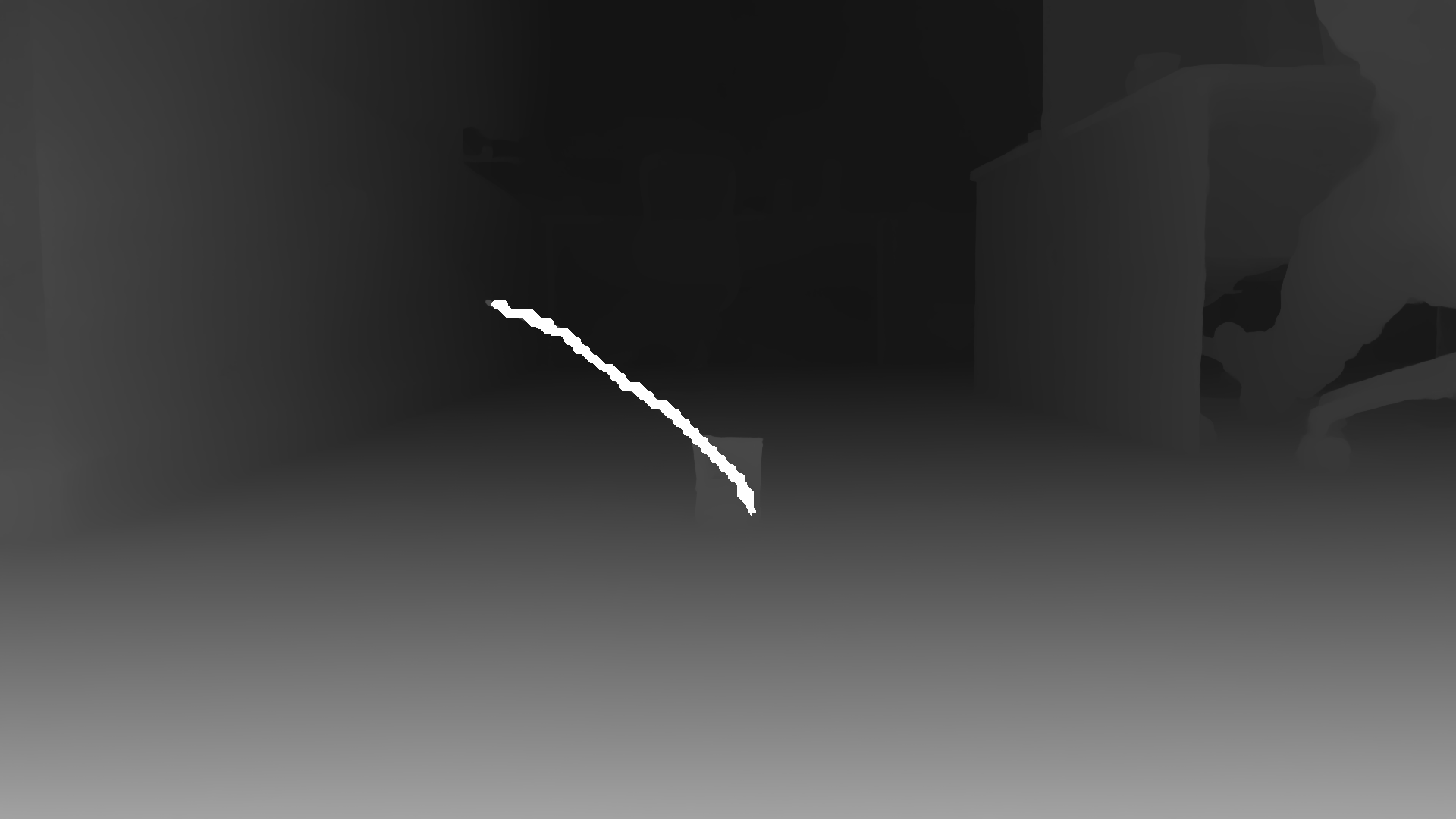}}
        \subfigure[NeRF-Stereo depth 2m]{\includegraphics[width=0.3\textwidth]{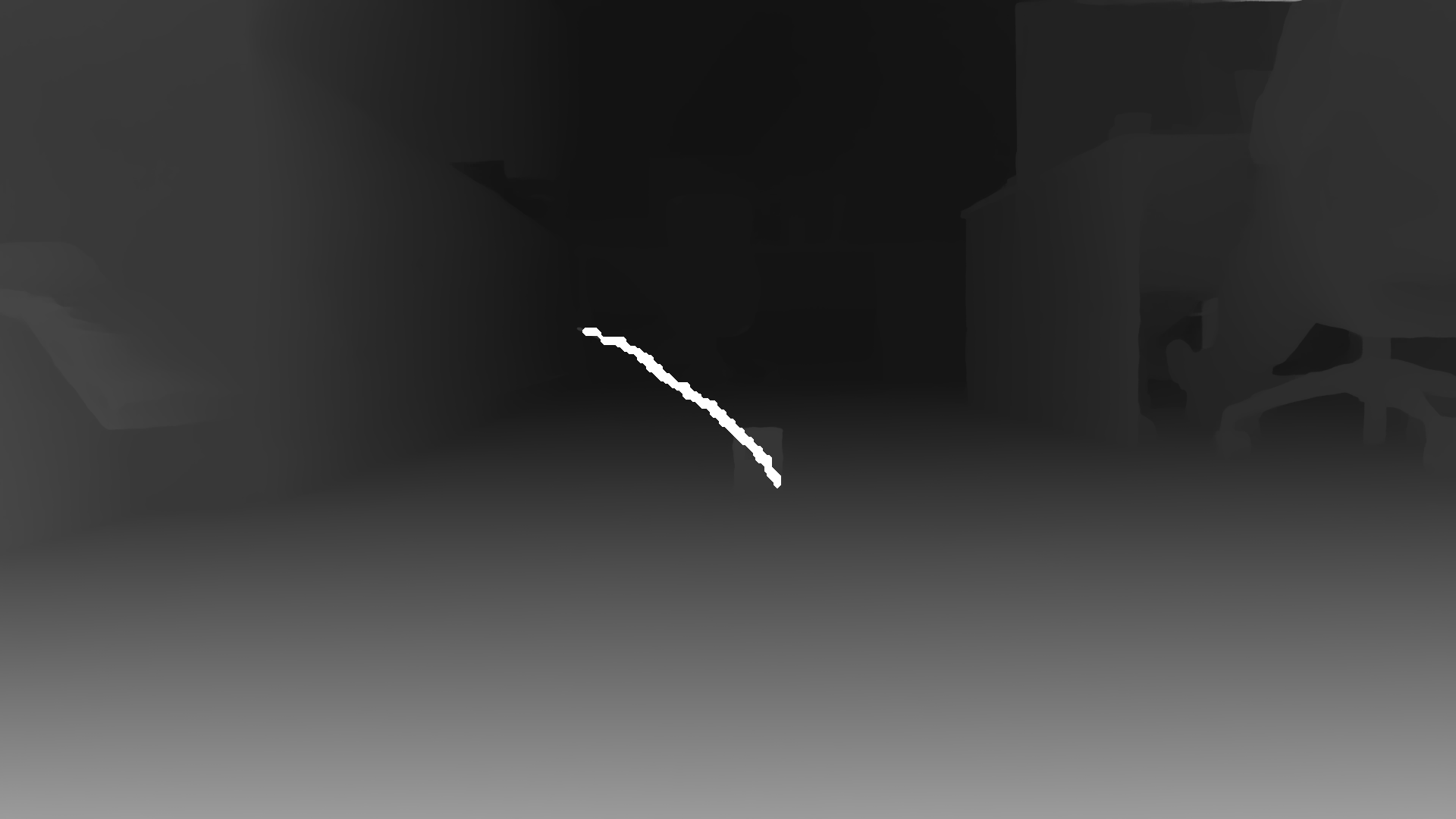}}
        \caption{ Comparison of Depth Estimation Using SGBM and NeRF at Varying Distances}
        \label{fig:depth_comparison}
    \end{figure}

    \begin{figure}[htbp]
        \centering
        \subfigure[SGBM histogram 1m]{\includegraphics[width=0.3\textwidth]{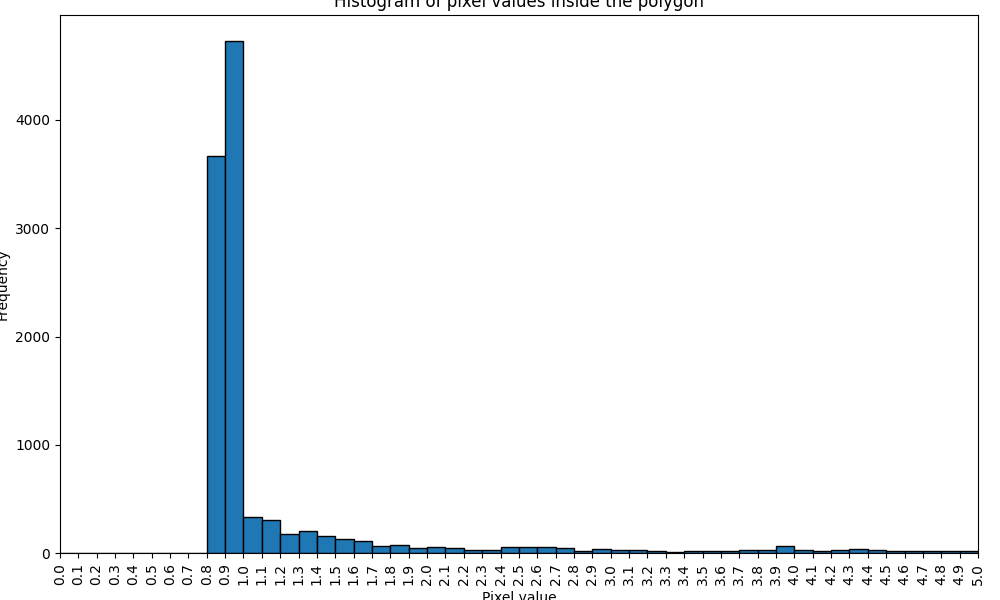}}
        \subfigure[SGBM histogram 1.5m]{\includegraphics[width=0.3\textwidth]{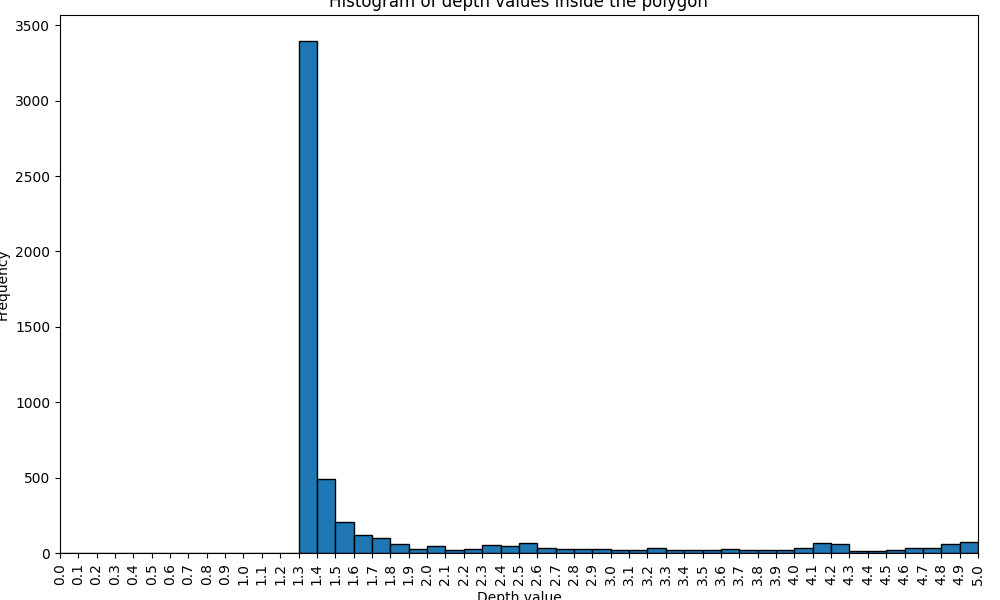}}
        \subfigure[SGBM histogram 2m]{\includegraphics[width=0.3\textwidth]{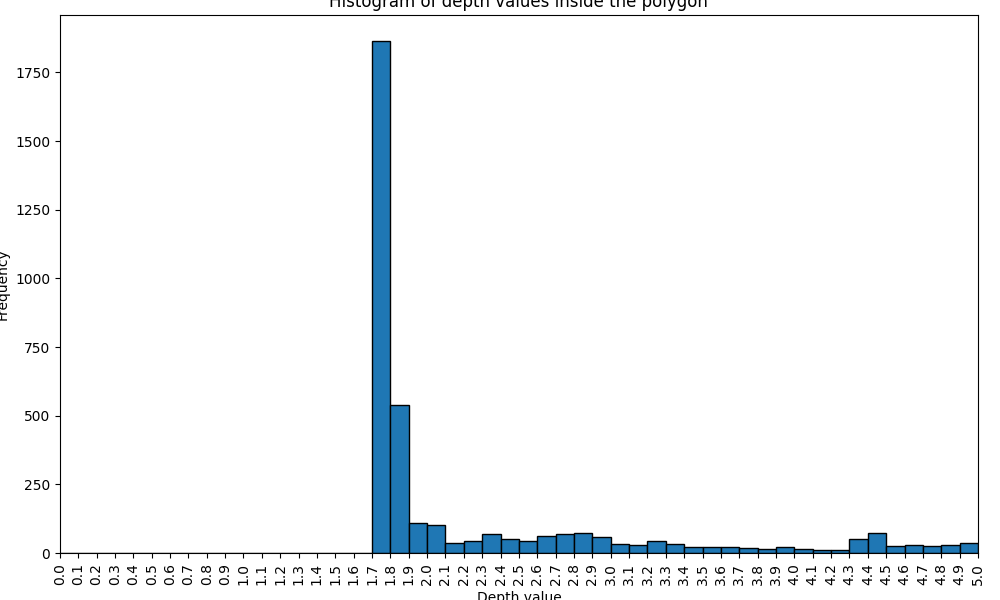}}
        \subfigure[NeRF-Stereo histogram 1m]{\includegraphics[width=0.3\textwidth]{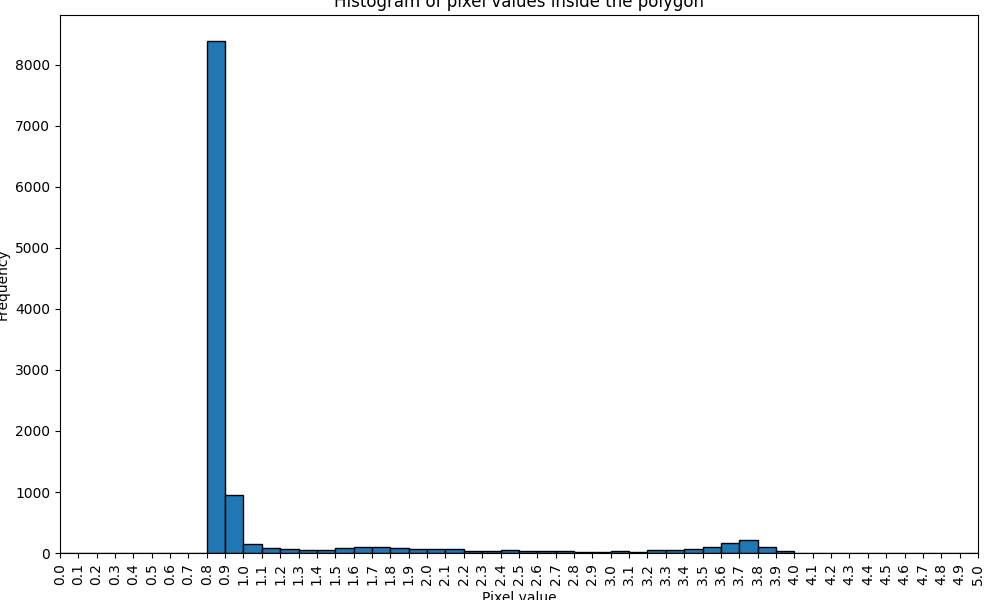}}
        \subfigure[NeRF-Stereo histogram 1.5m]{\includegraphics[width=0.3\textwidth]{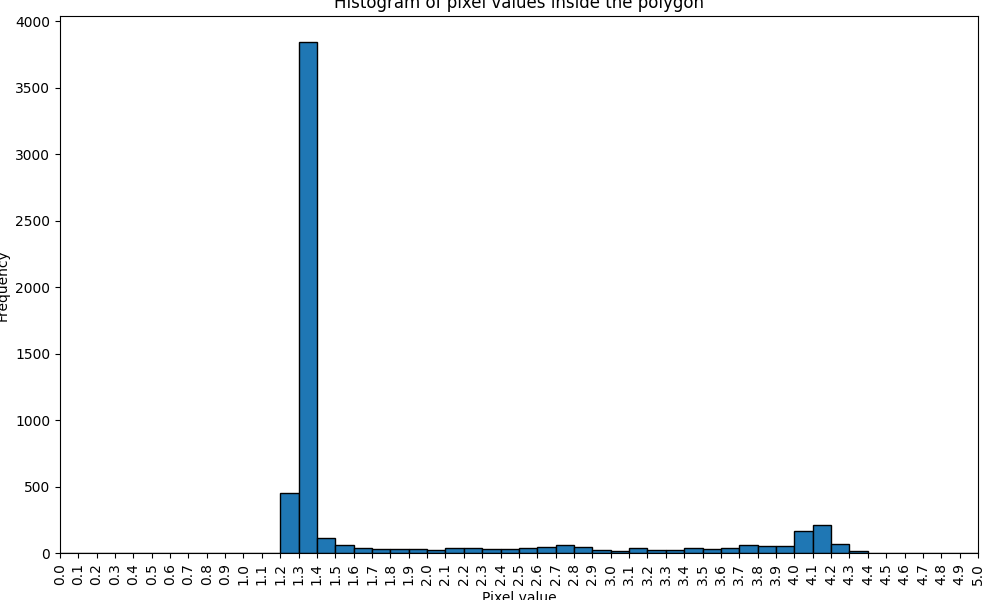}}
        \subfigure[NeRF-Stereo histogram 2m]{\includegraphics[width=0.3\textwidth]{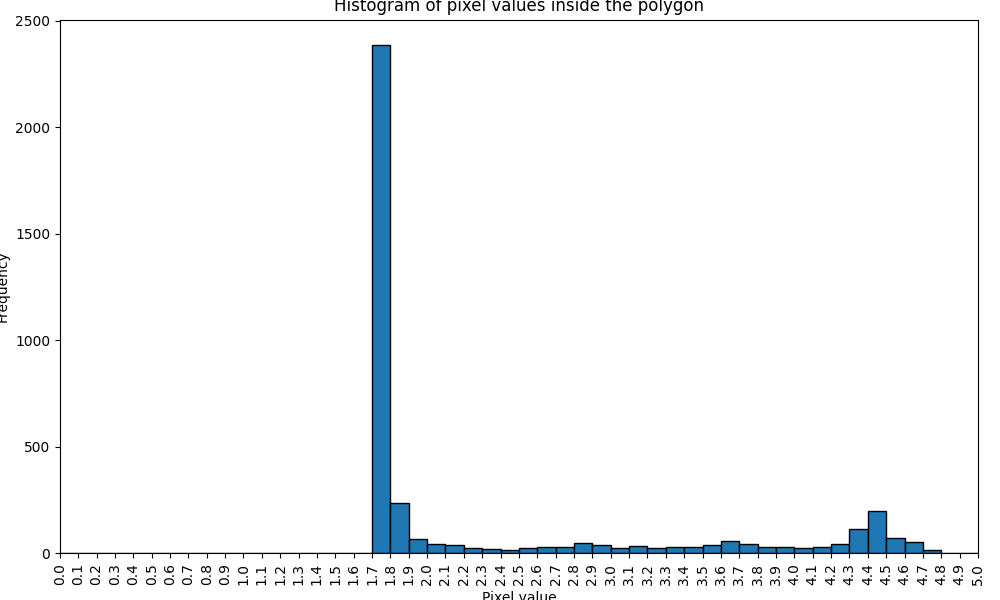}}
        \caption{Comparison of Depth Estimation Using SGBM and NeRF at Varying
        Distances}
        \label{fig:histograms}
    \end{figure}

    \section{Future Work}
    Future work encompasses three interrelated directions: expanding the data
    collection to include both indoor and outdoor environments, generating improved
    ground-truth depth maps through diverse datasets and methodologies, and
    exploring various neural network architectures to identify the most
    effective model. Completion of these efforts will bring the project close to
    its final stage.

    During data collection, images from a range of environments will be gathered,
    and multiple depth-generation techniques will be applied. Where feasible, LiDAR
    can supplement the stereo imagery, filling in missing data and enhancing
    depth-map precision.

    With high-quality ground-truth depth maps in hand, model training can proceed
    along two paths: one involves directly training a neural network
    architecture purpose-built for depth map estimation, while the other takes an
    indirect route through pre-existing deep architectures adapted to the task.
    The overall workflow is illustrated in the figure below.

    \begin{figure}[htbp]
        \centering
        \includegraphics[width=14cm]{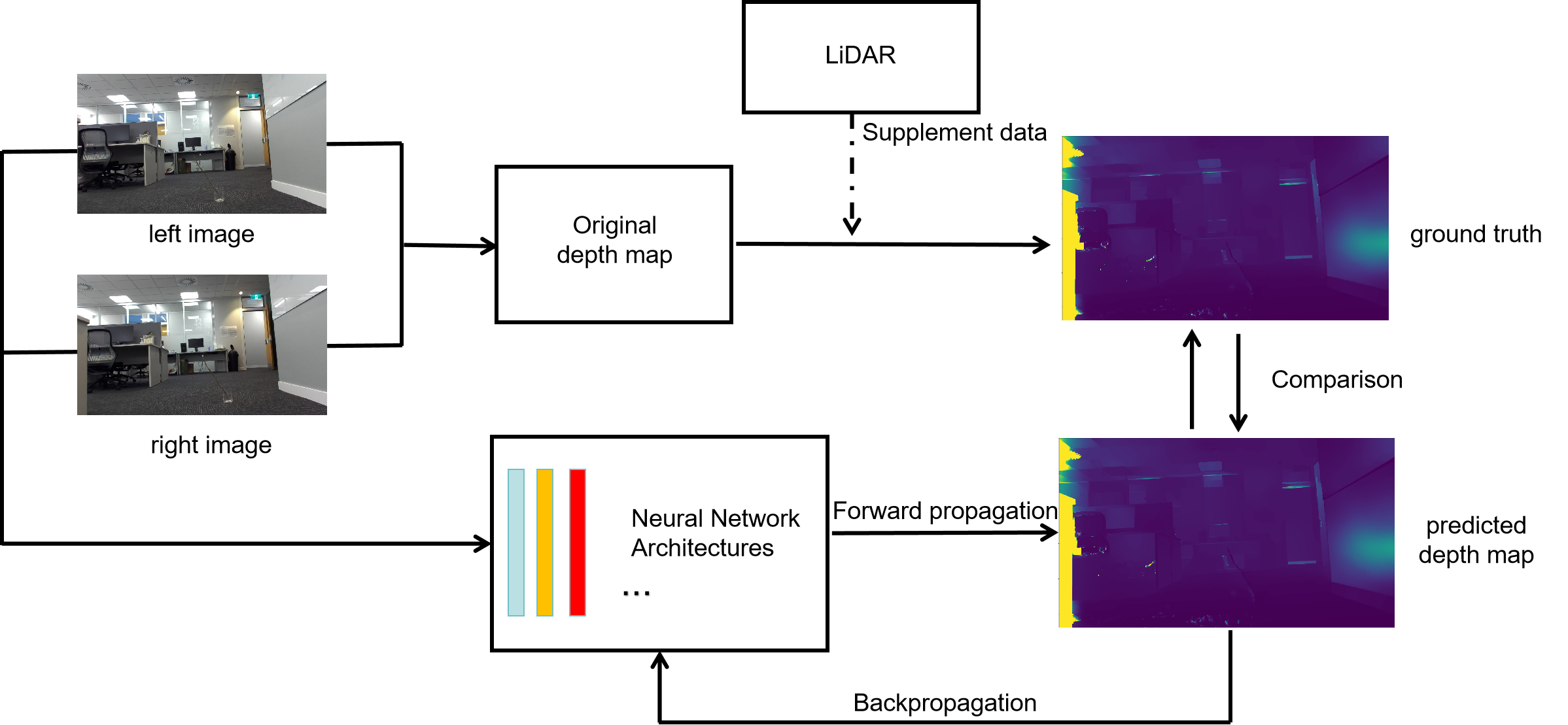}
        \caption{Training process of the depth map model}
        \label{fig:training_process}
    \end{figure}

    Numerous methodologies for training depth maps have been proposed, along
    with a wide variety of neural network architectures. For instance, the
    architecture presented in the Group-wise Correlation Stereo Network research,
    detailed in the Appendix, exemplifies one such approach. In the final stage of
    this project, various existing methods will be explored, and modifications to
    the network will be made by replacing certain components, such as
    experimenting with alternative activation functions beyond ReLU \citep{arora2016understanding}.
    Additionally, different model configurations will be trained and compared. A
    comprehensive evaluation will be conducted, with particular focus on key metrics
    such as prediction time and accuracy, to assess the performance of each
    model variant.

    \section{Conclusion}

    This paper examines the use of computer vision techniques for estimating tree
    branch depth, a key requirement for precision drone-assisted pruning. The research
    is structured into two primary components: the detection and segmentation of
    tree branches, and the generation of depth maps. Within the depth map
    generation process, this work provides a comprehensive analysis of traditional
    algorithms versus deep learning-based approaches. Traditional methods, which
    encompass procedures such as matching cost calculation, cost aggregation, and
    minimum cost selection, are straightforward to implement but limited in
    accuracy and generalisation. In contrast, deep learning methods use neural
    network architectures to capture finer features, producing higher-resolution
    and more accurate depth maps.

    By combining branch segmentation with the depth maps produced through deep learning,
    this work enables accurate measurement of branch distance, supporting more efficient
    and precise pruning operations with reduced risk of tree damage.

    The paper concludes by proposing several avenues for future research,
    including the refinement of deep learning models to improve real-time
    processing capabilities and accuracy, the development of more efficient resource
    management strategies to address the high computational demands of these models,
    and the extension of the method’s applicability across diverse environmental
    conditions and tree species. These recommendations provide valuable guidance
    for the advancement of automated pruning technologies and broader applications
    in agricultural and forestry automation.


    \bibliographystyle{plainnat}
    \bibliography{interactcsesample}
\end{document}